\documentclass[10pt,journal,compsoc]{IEEEtran}

\usepackage{algorithm}
\usepackage{algorithmicx}
\usepackage{algpseudocode}
\usepackage{amsmath}
\usepackage{diagbox}
\usepackage{graphics}
\usepackage{epsfig}
\usepackage{threeparttable}
\usepackage{xspace}
\usepackage{graphicx}
\usepackage{amssymb}
\usepackage{threeparttable}
\usepackage{color}
\usepackage[normalem]{ulem}
\usepackage{multirow}
\usepackage{float}
\usepackage{amsfonts}
\usepackage{bm}
\usepackage{array}
\usepackage[table]{xcolor}
\usepackage{colortbl}
\usepackage{diagbox}
\usepackage{rotating}
\usepackage{booktabs}
\usepackage{overpic}
\usepackage{contour}
\usepackage{enumitem}
\usepackage{cite}
\usepackage[pagebackref=false,breaklinks=true,colorlinks,bookmarks=false]{hyperref}

\definecolor{mygray}{gray}{.92}

\newcommand{\tabincell}[2]{\begin{tabular}{@{}#1@{}}#2\end{tabular}}

\makeatletter
\newcommand{\thickhline}{%
    \noalign {\ifnum 0=`}\fi \hrule height 1pt
    \futurelet \reserved@a \@xhline
}
\makeatother

\newcommand{\myfont}{\fontsize{7.9pt}{\baselineskip}\selectfont}

\makeatletter
\DeclareRobustCommand\onedot{\futurelet\@let@token\@onedot}
\def\@onedot{\ifx\@let@token.\else.\null\fi\xspace}

\def\eg{\emph{e.g}\onedot} 
\def\ie{\emph{i.e}\onedot} 
 
\def\etc{\emph{etc}\onedot} 
\def\wrt{w.r.t\onedot} 
\def\etal{\emph{et al}\onedot}

\def\cred{\leavevmode\color{red}}
\def\cblue{\leavevmode\color{blue}}
\makeatother

\begin{document}
\title{Salient Object Detection in the \\Deep Learning Era: An In-depth Survey}

\author{Wenguan~Wang,~\IEEEmembership{Member,~IEEE}, Qiuxia Lai, Huazhu Fu,~\IEEEmembership{Senior Member,~IEEE}, \\
Jianbing~Shen,~\IEEEmembership{Senior Member,~IEEE}, Haibin Ling, and Ruigang Yang,~\IEEEmembership{Senior Member,~IEEE}
\IEEEcompsocitemizethanks{
\IEEEcompsocthanksitem W. Wang is  with ETH Zurich, Switzerland.
(Email: wenguanwang.ai@gmail.com)
\IEEEcompsocthanksitem Q. Lai is with the Department of Computer Science and
Engineering, the Chinese University of Hong Kong, Hong Kong, China. (Email: qxlai@cse.cuhk.edu.hk)
\IEEEcompsocthanksitem H. Fu is with Inception Institute of Artificial Intelligence, UAE.
(Email: hzfu@ieee.org)
\IEEEcompsocthanksitem J. Shen is with Beijing Laboratory of Intelligent Information Technology,
School of Computer Science, Beijing Institute of Technology, China.
(Email: shenjianbing@bit.edu.cn)
\IEEEcompsocthanksitem H. Ling is with the Department of Computer and Information Sciences,
Temple University, Philadelphia, PA, USA. (Email: hbling@temple.edu)
\IEEEcompsocthanksitem R. Yang is with the University of Kentucky, Lexington, KY 40507. (Email:
ryang@cs.uky.edu)
\IEEEcompsocthanksitem Corresponding author: \textit{Jianbing Shen}
%
}
}

\markboth{IEEE TRANSACTIONS ON PATTERN ANALYSIS AND MACHINE INTELLIGENCE}%
{Shell \MakeLowercase{\textit{et al.}}: Bare Demo of IEEEtran.cls for Journals}

\IEEEtitleabstractindextext{
\begin{abstract}
As an essential problem in computer vision, salient object detection (SOD) has attracted an increasing amount of research {attention} over the years. Recent advances in SOD 
are {predominantly} led by deep learning-based solutions (named deep {SOD}).
To {enable} in-depth understanding of deep {SOD}, in this paper, we provide a comprehensive survey covering various aspects, ranging from algorithm taxonomy to unsolved issues. In particular, we first review deep SOD algorithms from different perspectives, including network architecture, level of supervision, learning paradigm, and {object-/instance-}level detection. Following that, we summarize and analyze existing SOD datasets and evaluation metrics. Then, we benchmark a large group of representative SOD models, and provide detailed analyses of the comparison results. Moreover, we study the performance of SOD algorithms under different {attribute settings}, which {has not been thoroughly} explored previously, by constructing a novel SOD dataset with rich attribute annotations
{covering various salient object types, challenging factors, and scene categories.}
We further analyze, for the first time in the field, the robustness of SOD models {to} random input perturbations and adversarial attacks.
We also look into the generalization and {difficulty} of existing SOD datasets. Finally, we discuss several open issues of SOD and outline future research directions.
All the saliency prediction maps, our constructed dataset with annotations, and codes for evaluation are publicly available at \url{https://github.com/wenguanwang/SODsurvey}.
\end{abstract}
}
\maketitle

\IEEEdisplaynontitleabstractindextext
\IEEEpeerreviewmaketitle

\IEEEraisesectionheading{\section{Introduction}\label{sec:introduction}}
\IEEEPARstart{S}{alient} object detection (SOD) aims at {highlighting visually salient object regions in images. Here, `visually salient' describes the property of an object or a region to attract human observers' attention. SOD is driven by and applied to a wide spectrum of \textit{object-level} applications in various areas. In computer vision, representative applications include
image understanding~\cite{zhu2015unsupervised,zhang2015saliency},
image captioning~\cite{xu2015show,fang2015captions,das2017human},
object detection~\cite{ren2014region,zhang2016bridging},
{unsupervised} video object segmentation~\cite{wang2018saliency,Song_2018_ECCV},
semantic segmentation~\cite{wei2017object,Wang_2018_CVPRWeakly,sun2020mining},
person re-identification~\cite{zhao2013unsupervised,bi2014person},
{and} video summarization~\cite{ma2002user,simakov2008summarizing}.
In computer graphics, SOD {also} plays an essential role in various tasks, {including}
{non-photorealistic} rendering\!~\cite{han2013fast,rosin2013artistic},
image cropping\!~\cite{wang2016stereoscopic,wang2018deepcrop},
image retargeting~\cite{avidan2007seam}, \etc.
{Several} applications in robotics, {such as}
human-robot interaction~\cite{sugano2010calibration,borji2014defending}, and
object discovery~\cite{karpathy2013object,frintrop2014cognitive},
also benefit from SOD for better scene/object understanding.

{Though inspired by eye} fixation prediction (FP)~\cite{treisman1980feature}, which originated from cognitive and psychology research communities to investigate the human attention mechanism by predicting eye fixation positions in visual scenes, SOD {differs in that it} aims {to detect} the whole attentive object regions. {Since} the renaissance of deep learning techniques, significant improvement for SOD has been {achieved} in recent years, {thanks to} the powerful representation learning methods. Since the first introduction in 2015~\!\cite{li2015visual,wang2015deep,zhao2015saliency}, deep learning-based SOD (or \emph{deep SOD}) algorithms have {quickly} shown superior performance over traditional solutions, and {have continued to improve the state-of-the-art}.

This paper provides a comprehensive and in-depth survey of SOD in the deep learning era. In addition to taxonomically reviewing existing deep SOD methods, it provides in-depth analyses of representative datasets and evaluation metrics, {and} investigates crucial but largely under-explored issues, such as {the} robustness and transferability of deep SOD models, {their} strengths
and weaknesses {under} certain scenarios (\ie, scene/salient object categories, challenging factors), {as well as the} generalizability and {difficulty} of SOD datasets. The saliency maps used for benchmarking, our constructed dataset, and evaluation codes are available at \url{https://github.com/wenguanwang/SODsurvey}.

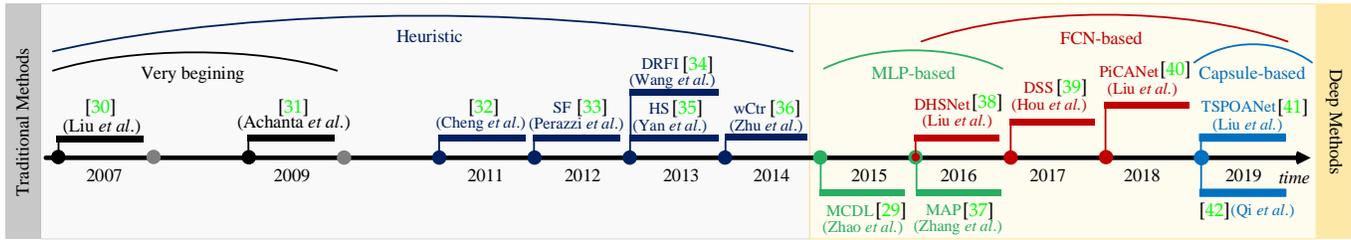
\begin{figure*}[t]
  \centering
      \includegraphics[width=0.99 \linewidth]{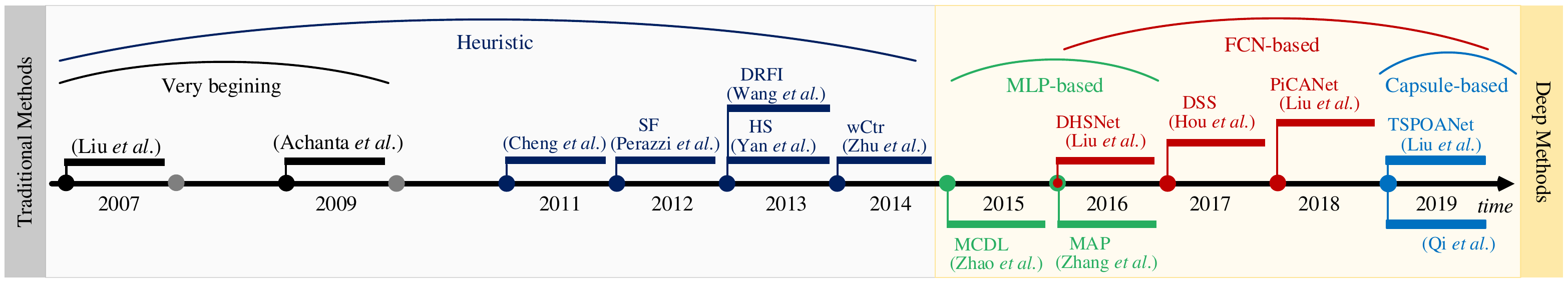}
      \put(-483,50){\scriptsize{\cite{Liu2007}}}
      \put(-410,50){\scriptsize{\cite{achanta2009frequency}}}
      \put(-338,50){\scriptsize{\cite{cheng2011global}}}
      \put(-297,50){\scriptsize{\cite{perazzi2012saliency}}}
      \put(-257,66){\scriptsize{\cite{WangDRFI2017}}}
      \put(-261,49.5){\scriptsize{\cite{yan2013hierarchical}}}
      \put(-224,49.5){\scriptsize{\cite{zhu2014saliency}}}
      \put(-184,10.8){\scriptsize{\cite{zhao2015saliency}}}
      \put(-151,10.8){\scriptsize{\cite{zhang2016unconstrained}}}
      \put(-147.6,51.6){\scriptsize{\cite{liu2016dhsnet}}}
      \put(-115.6,57.2){\scriptsize{\cite{hou2016deeply}}}
      \put(-76.4,64){\scriptsize{\cite{liu2018picanet}}}
      \put(-32,50){\scriptsize{\cite{liu2019employing}}}
      \put(-61.6,10.8){\scriptsize{\cite{qi2019multi}}}
\caption{A brief chronology of SOD. The very first SOD models date back to the work of Liu \etal~\cite{Liu2007} and Achanta \etal~\cite{achanta2009frequency}. The first incorporation of deep learning techniques {into} SOD models {was in} $2015$. Listed methods are milestones, which are typically highly cited. See \S\ref{sec:oah} for more details.
}
\label{fig:chronology}
\end{figure*}

\begin{table*}[!t]
	\centering
	\caption{Summary of previous reviews. {For each work, the publication information and coverage are provided.} See \S\ref{sec:cmp_rv} for more detailed descriptions.}
	\vspace{-5pt}
\renewcommand\arraystretch{1.05}
	\begin{threeparttable}
		\resizebox{0.99\textwidth}{!}{
			\setlength\tabcolsep{2pt}
      \begin{tabular}{|l|c|c|l|}
				\hline\thickhline
        Title &Year &Venue &Description \\
				\hline
				\hline
        State-of-the-Art in Visual Attention Modeling~\cite{borji2013state} &2013 &TPAMI
				&This paper reviews visual attention (\ie fixation prediction) models before 2013. \\
				\hline
        Salient Object Detection: A Benchmark~\cite{borji2015salient} &2015 &TIP
				&This paper benchmarks 29 heuristic SOD models and 10 FP methods over 7 datasets. \\
				\hline
        Attentive Systems: A Survey~\cite{nguyen2018attentive} &2017 &IJCV
				&This paper reviews applications that utilize visual saliency cues. \\
				\hline
        \tabincell{l}{A Review of Co-Saliency Detection Algorithms:  \\
					Fundamentals, Applications, and Challenges~\cite{zhang2018review}} &2018 &TIST
				&This paper reviews the fundamentals, challenges, and applications of co-saliency detection.\\
				\hline
        \tabincell{l}{Review of Visual Saliency Detection with Comprehen- \\
					sive Information~\cite{cong2018review}} &2018 &TCSVT
				&This paper reviews RGB-D SOD, co-saliency detection and video SOD. \\
				\hline
        \tabincell{l}{Advanced Deep-Learning Techniques for Salient and \\
					Category-Specific Object Detection: A Survey~\cite{han2018advanced}} &2018 &SPM
				&\tabincell{l}{This paper reviews several sub-directions of object detection, namely objectness detection, SOD\\ and category-specific object detection.} \\
				\hline
        \tabincell{l}{Saliency Prediction in the Deep Learning Era: Successes\\
				 and Limitations~\cite{borji2019saliency}} &2019 &TPAMI
				&This paper reviews image and video fixation prediction models and analyzes specific questions. \\
				\hline
        Salient Object Detection: A Survey~\cite{borji2014salient} &2019 &CVM
				&\tabincell{l}{This paper reviews 65 heuristic and 21 deep SOD models {up to} 2017 and discusses  closely related \\areas like object detection, fixation prediction, segmentation, \etc.} \\
				\hline
				
			\end{tabular}
		}
	\end{threeparttable}
	\label{table:p_reviews}
\end{table*}

\subsection{History and Scope}\label{sec:oah}
Humans are able to quickly allocate attention to important regions in visual scenes. Understanding and modeling such an astonishing ability, \ie, visual attention or visual saliency, is a fundamental research problem in psychology, neurobiology, cognitive science, and computer vision. There are two categories of computational models for visual saliency, namely FP and SOD. FP originated from cognitive and psychology communities~\cite{treisman1980feature,koch1985shifts,itti1998model}, {and targets at predicting \textit{where people look in images}.}

The history of SOD is relatively short and can be traced back to~\cite{Liu2007} and~\cite{achanta2009frequency}. The rise of SOD has been driven by a wide range of object-level computer vision applications. {Instead of FP models only predicting sparse eye fixation locations, SOD models aim to detect the whole entities of the visually attractive objects with precise boundaries.} Most traditional, \textbf{non-deep SOD models}~\cite{jiang2013salient,zhu2014saliency} rely on  low-level features and certain heuristics (\eg, \emph{color contrast}~\cite{cheng2011global}, \emph{background prior}~\cite{wei2012geodesic}). To obtain uniformly highlighted salient objects and clear object boundaries, an over-segmentation process that generates regions~\cite{shi2015hierarchical}, super-pixels~\cite{yang2013,wang2016correspondence}, or object proposals~\cite{guo2017video} is often integrated into these models. Please see~\cite{borji2015salient} for a more comprehensive overview.

With the compelling success of deep learning technologies in computer vision, more and more \textbf{deep SOD methods} have {begun} springing up since $2015$.
Earlier deep SOD models {utilized} multi-layer perceptron (MLP) classifiers to predict the saliency score of deep features extracted from each image processing unit~\cite{zhao2015saliency,wang2015deep,li2015visual}. Later, a more effective and efficient form, \ie, fully convolutional network (FCN)-based {model}, {became} the mainstream SOD architecture. Some recent {methods}~\cite{qi2019multi,liu2019employing} {also introduced} Capsule~\cite{hinton2011transforming} into SOD {to comprehensively address} object property modeling. A brief chronology of SOD is shown in Fig.~\ref{fig:chronology}.

\vspace{1mm}\noindent\textbf{Scope of the survey}.
Despite {its} short history, research in deep SOD has produced hundreds of papers, making it impractical (and fortunately unnecessary) to review all of them. Instead, we comprehensively select influential papers published in prestigious journals and conferences. This survey mainly focuses on the major progress in the last five years, but for completeness and better readability, some early related works are also included. Due to limitations on space and our knowledge, we apologize to those authors whose works are not included in this paper. It is worth noting that we restrict this survey to \textit{single-image SOD} methods, and leave RGB-D SOD, co-saliency detection, video SOD, \etc, as separate topics.

\subsection{Related Previous Reviews and Surveys}
\label{sec:cmp_rv}
Table~\ref{table:p_reviews} lists existing surveys that are related to ours.
Among them, Borji~\etal~\cite{borji2015salient} reviewed SOD methods preceding 2015, thus do not refer to recent deep learning-based solutions. Zhang~\etal~\cite{zhang2018review} reviewed methods for co-saliency detection, \ie, detecting common salient objects from multiple relevant images.
Cong~\etal~\cite{cong2018review} reviewed several extended SOD tasks including RGB-D SOD, co-saliency detection and video SOD. Han~\etal~\cite{han2018advanced} looked into several sub-directions of object detection, and {outlined} recent progress in objectness detection, SOD, and category-specific object detection. Borji~\etal summarized both heuristic~\cite{borji2013state} and deep models~\cite{borji2019saliency} for FP. Nguyen~\etal~\cite{nguyen2018attentive} focused on categorizing the applications of visual saliency (including both SOD and FP) in different areas.
{Finally, a more} recently published survey~\cite{borji2014salient} covers both traditional non-deep SOD methods and deep ones {until} 2017, and discusses {their} relation {to} several other closely-related research areas, such as special-purpose object detection and segmentation.

\begin{table*}[t]
\centering
\caption{{Taxonomies and representative publications of deep SOD methods. See \S\ref{sec:sod_models} for more detailed descriptions.}}
\begin{threeparttable}
\resizebox{1\textwidth}{!}{
\setlength\tabcolsep{2pt}
\renewcommand\arraystretch{1.02}
\begin{tabular}{|l l l|| l|}
\hline\thickhline
\multicolumn{3}{|c||}{Category} &\multicolumn{1}{c|}{Publications}\\
\hline
\hline
\multirow{9}{*}{\tabincell{l}{Network\\Architectures\\(\S\ref{sec:sod_network_structure})}}
  &\multirow{1}{*}{\textit{Multi-layer perceptron}}
    &1) Super-pixel/patch-based &\cite{zhao2015saliency}, \cite{lee2016deep}, \cite{li2015visual}, \cite{he2015supercnn} \\
    &(MLP)-based&2) Object proposal based   &\cite{wang2015deep}, \cite{zhang2016unconstrained}, \cite{kim2016shape} \\
  \cline{2-4}
  &\multirow{5}{*}{\tabincell{l}{\textit{Fully convolutional network} \\ (FCN)-based}}
    &1) Single-stream      &\cite{wang2016saliency}, \cite{kuen2016recurrent}, \cite{hu2017deep}, \cite{Zhang_2017_ICCV}, \cite{zhang2018deep}, \cite{cao2018lateral}, \cite{li2019supervae} \\
    &&2) Multi-stream       &\cite{li2017instance}, \cite{Wang_2017_ICCV}, \cite{chen2017look}, \cite{zeng2019towards}, \cite{zhuge2019deep} \\
    &&3) Side-fusion        &\cite{hou2016deeply}, \cite{luo2017non}, \cite{zhang2017amulet}, \cite{he2017delving}, \cite{hu2018recurrently}, \cite{Islam_2018_CVPR}, \cite{wu2019cascaded}, \cite{zeng2019multi}, \cite{zhao2019egnet} \\
    &&4) Bottom-up/top-down &\cite{liu2016dhsnet}, \cite{zhang2017supervision}, \cite{zhang2018bi}, \cite{wang2018detect}, \cite{zhang2018progressive}, \cite{Wang_2018_CVPR}, \cite{liu2018picanet}, \cite{chen2018reverse}, \cite{feng2019attentive}, \cite{qin2019basnet}, \cite{wu2019mutual}, \cite{wang2019salient}, \cite{liu2019simple}, \cite{wang2019iterative}, \cite{xu2019structured} \\
    &&5) Branched           &\cite{kruthiventi2016saliency}, \cite{wang2017}, \cite{li2018weakly}, \cite{li2018contour}, \cite{zhang2019capsal}, \cite{su2019selectivity}, \cite{wu2019stacked}, \cite{zeng2019joint} \\
  \cline{2-4}
  &\textit{Hybrid network}-based & &\cite{li2016deep}, \cite{tang2016saliency} \\
  \cline{2-4}
  &\textit{Capsule}-based & &\cite{liu2019employing}, \cite{qi2019multi} \\
\hline

\multirow{3}{*}{\tabincell{l}{Level of\\Supervision\\(\S\ref{sec:sod_supervision})}}
  &\textit{Fully-supervised} & &\textit{All others} \\
  \cline{2-4}
  &\multirow{2}{*}{\textit{Un-/Weakly-supervised}}
    &1) Category-level     &\cite{wang2017}, \cite{cao2018lateral}, \cite{li2019supervae}, \cite{zeng2019multi} \\
    &&2) Pseudo pixel-level &\cite{zhang2017supervision}, \cite{li2018weakly}, \cite{zhang2018deep}, \cite{li2018contour} \\
\hline

\multirow{7}{*}{\tabincell{l}{Learning\\Paradigm\\(\S\ref{sec:sod_learning_paradigm})}}
  &\textit{Single-task learning} (STL) & &\textit{All others} \\
  \cline{2-4}
  &\multirow{6}{*}{\tabincell{l}{\textit{Mingle-task learning}\\\textit{(MTL)}}}
    &1) Salient object subitizing &\cite{zhang2016unconstrained}, \cite{he2017delving}, \cite{Islam_2018_CVPR} \\
    &&2) Fixation prediction       &\cite{kruthiventi2016saliency}, \cite{Wang_2018_CVPR} \\
    &&3) Image classification      &\cite{wang2017}, \cite{li2018weakly} \\
    &&4) Semantic segmentation     &\cite{wang2016saliency}, \cite{zeng2019joint} \\
    &&5) Contour/edge detection    &\cite{luo2017non}, \cite{li2018contour}, \cite{feng2019attentive}, \cite{wu2019mutual}, \cite{wang2019salient}, \cite{liu2019simple}, \cite{su2019selectivity}, \cite{zhao2019egnet}, \cite{wu2019stacked} \\
    &&6) Image captioning          &\cite{zhang2019capsal} \\
\hline

\multirow{2}{*}{\tabincell{l}{Object-/Instance-\\Level (\S\ref{sec:sod_pixel_instance})}}
  &\textit{Object-level} & &\textit{All others} \\
  \cline{2-4}
  &\textit{Instance-level} & &\cite{zhang2016unconstrained}, \cite{li2017instance} \\
\hline
\end{tabular}
}
\end{threeparttable}
\label{table:sod_texomonies}
\vspace{-8pt}
\end{table*}

Different from previous SOD surveys, {which} focus on earlier non-deep learning SOD methods~\cite{borji2015salient}, other related fields~\cite{cong2018review,han2018advanced,borji2013state,borji2019saliency}, practical applications~\cite{nguyen2018attentive} or {a} limited number of deep SOD models~\cite{borji2014salient}, this work systematically and comprehensively reviews recent advances in the field. It {features} in-depth {analyses and discussions on} various aspects, many of which, to the best of our knowledge, {have never been explored} in this field. In particular, we {comprehensively} summarize and {discuss} existing deep SOD methods {under} several proposed taxonomies (\S\ref{sec:sod_models}){;} review datasets (\S\ref{sec:soddatasets}) and evaluation metrics (\S\ref{sec:evametric}) with their pros and cons{;} {provide} a deeper understanding of SOD models through {an} attribute-based evaluation (\S\ref{sec:attribute_eval}){;} discuss the influence of input perturbation (\S\ref{sec:input_ptb}){;} analyze the robustness of deep SOD models {to} adversarial attacks (\S\ref{sec:adversarial_atk}){;} study the generalization and {difficulty} of existing SOD datasets (\S\ref{sec:cd_generalization}){;} and offer insight {into} essential open issues, challenges, and future directions (\S\ref{sec:discussions}).
We expect our survey to provide novel insight and inspiration {that will facilitate} the understanding of deep SOD, and foster research on {the open issues raised}.

\subsection{Our Contributions}\label{sub:contribution}
Our contributions in this paper are summarized as follows:
\begin{enumerate}[leftmargin=*]
\item \textbf{A systematic review of deep SOD models from various perspectives.}
    We categorize and summarize existing deep SOD models according to network architecture, level of supervision, learning paradigm, \etc. The proposed taxonomies aim to help researchers {gain} a deeper understanding of the key features of deep SOD models.
\item \textbf{An attribute-based performance evaluation of SOD models.}
    We compile a hybrid dataset and provide annotated attributes {for} object categories, scene categories, and challenging factors. By evaluating several representative SOD models on it, {we} uncover {the} strengths and weaknesses of deep and non-deep approaches, opening up promising directions for future efforts.
\item \textbf{An analysis of the robustness of SOD models against general input perturbations.} To study the robustness of SOD models, we investigate the effects of various perturbations on {the} final performance of deep and non-deep SOD models. Some results are somewhat unexpected.
\item \textbf{The first known adversarial attack analysis {for} SOD models.}
    We further examine {the} robustness of SOD models against intentionally designed perturbations, \ie, adversarial attacks. The specially designed attacks and evaluations can serve as baselines for {further studying} the robustness and transferability of deep SOD models.
\item \textbf{Cross-dataset generalization study.}
    To analyze the generalization and difficulty of existing SOD datasets in-depth, we conduct a cross-dataset generalization study that quantitatively reveals the dataset bias.
\item \textbf{Overview of open issues and future directions.}
    We thoroughly look over several essential issues (\ie, model design, dataset collection, \etc), {shedding} light on potential directions for future research.
\end{enumerate}
These contributions {together comprise} an exhaustive, up-to-date, and in-depth survey, and differentiate it from previous review papers significantly.

\begin{table*}
\centering
\caption{
        {Summary of essential characteristics for popular SOD methods.
        Here, `\#Training' is the number of training images, and `CRF' denotes whether the predictions are post-processed by conditional random field~\cite{krahenbuhl2011efficient}. See \S\ref{sec:sod_models} for more detailed descriptions.}
        }
\label{table:sod_methods}
\begin{threeparttable}
\resizebox{1\textwidth}{!}{
\setlength\tabcolsep{2pt}
\renewcommand\arraystretch{1.02}
\begin{tabular}{|c|r||c|c|c|c|c|c|c|c|c|}
\hline\thickhline

&Methods~~~~  &Publ. &Architecture &Backbone
    &\tabincell{c}{Level of\\Supervision} &\tabincell{c}{Learning\\Paradigm} &\tabincell{c}{Obj.-/Inst.-\\Level SOD} &Training Dataset &\#Training &CRF\\
\hline
\hline
\multirow{4}{*}{\rotatebox{90}{2015}}
&SuperCNN~\cite{he2015supercnn} &IJCV &MLP+super-pixel &-
    &Fully-Sup. &STL &Object &ECSSD~\cite{shi2015hierarchical} &800&\\
&MCDL~\cite{zhao2015saliency}  &CVPR &MLP+super-pixel &GoogleNet 
    &Fully-Sup. &STL &Object &MSRA10K~\cite{ChengPAMI} &8,000&\\

&LEGS~\cite{wang2015deep}  &CVPR &MLP+segment &-
    &Fully-Sup. &STL &Object &MSRA-B~\cite{Liu2007}+PASCAL-S~\cite{li2014secrets} &3,000+340 &\\

&MDF~\cite{li2015visual} &CVPR &MLP+segment &-
    &Fully-Sup. &STL &Object &MSRA-B~\cite{Liu2007} &2,500&\\
\hline
\multirow{9}{*}{\rotatebox{90}{2016}}
&ELD~\cite{lee2016deep} &CVPR &MLP+super-pixel &VGGNet
    &Fully-Sup. &STL &Object &MSRA10K~\cite{ChengPAMI} &$\sim$9,000&\\
&DHSNet~\cite{liu2016dhsnet} &CVPR &FCN &VGGNet
    &Fully-Sup. &STL &Object &MSRA10K~\cite{ChengPAMI}+DUT-OMRON~\cite{yang2013} &6,000+3,500&\\
&DCL~\cite{li2016deep} &CVPR &FCN &VGGNet
    &Fully-Sup. &STL &Object &MSRA-B~\cite{Liu2007} &2,500 &\checkmark\\
&RACDNN~\cite{kuen2016recurrent} &CVPR  &FCN &VGGNet
    &Fully-Sup. &STL &Object &DUT-OMRON~\cite{yang2013}+NJU2000~\cite{ju2015depth}+RGBD~\cite{peng2014rgbd} &10,565&\\
&SU~\cite{kruthiventi2016saliency} &CVPR  &FCN &VGGNet
 &Fully-Sup. &MTL &Object &MSRA10K~\cite{ChengPAMI}+SALICON~\cite{jiang2015salicon} &10,000+15,000 &\checkmark\\
&MAP~\cite{zhang2016unconstrained} &CVPR  &MLP+obj. prop. &VGGNet
 &Fully-Sup. &MTL &Instance &SOS~\cite{zhang2015salient} &$\sim$5,500&\\
&SSD~\cite{kim2016shape} &ECCV &MLP+obj. prop. &AlexNet
    &Fully-Sup. &STL &Object &MSRA-B~\cite{Liu2007} &2,500&\\
&CRPSD~\cite{tang2016saliency} &ECCV &FCN &VGGNet
    &Fully-Sup. &STL &Object &MSRA10K~\cite{ChengPAMI} &10,000&\\
&RFCN~\cite{wang2016saliency} &ECCV &FCN &VGGNet
    &Fully-Sup. &MTL &Object &PASCAL VOC 2010~\cite{everingham2010pascal}+MSRA10K~\cite{ChengPAMI} &10,103+10,000&\\
\hline
\multirow{11}{*}{\rotatebox{90}{2017}}
&MSRNet~\cite{li2017instance} &CVPR &FCN &VGGNet
    &Fully-Sup. &STL &Instance &MSRA-B~\cite{Liu2007}+HKU-IS~\cite{li2015visual}  (+ILSO~\cite{li2017instance}) &2,500+2,500 (+500) &\checkmark\\
&DSS~\cite{hou2016deeply}  &CVPR &FCN &VGGNet
    &Fully-Sup. &STL &Object &MSRA-B~\cite{Liu2007}+HKU-IS~\cite{li2015visual} &2,500 &\checkmark\\
&WSS~\cite{wang2017}  &CVPR &FCN &VGGNet
    &Weakly-Sup. &MTL&Object &ImageNet~\cite{deng2009imagenet} &456k &\checkmark\\
&DLS~\cite{hu2017deep} &CVPR &FCN &VGGNet
    &Fully-Sup. &STL &Object &MSRA10K~\cite{ChengPAMI} &10,000&\\
&NLDF~\cite{luo2017non} &CVPR &FCN &VGGNet
    &Fully-Sup. &MTL &Object &MSRA-B~\cite{Liu2007} &2,500 &\checkmark\\
&DSOS~\cite{he2017delving} &ICCV &FCN &VGGNet
    &Fully-Sup. &MTL &Object &SOS~\cite{zhang2015salient} &6,900&\\
&Amulet~\cite{zhang2017amulet} &ICCV &FCN &VGGNet
    &Fully-Sup. &STL &Object &MSRA10K~\cite{ChengPAMI} &10,000&\\
&FSN~\cite{chen2017look}  &ICCV &FCN &VGGNet
    &Fully-Sup. &STL &Object &MSRA10K~\cite{ChengPAMI} &10,000&\\
&SBF~\cite{zhang2017supervision} &ICCV &FCN &VGGNet
    &Un-Sup. &STL &Object &MSRA10K~\cite{ChengPAMI} &10,000&\\
&SRM~\cite{Wang_2017_ICCV}  &ICCV &FCN &ResNet
    &Fully-Sup. &STL &Object &DUTS~\cite{wang2017} &10,553&\\
&UCF~\cite{Zhang_2017_ICCV}  &ICCV &FCN &VGGNet
    &Fully-Sup. &STL &Object &MSRA10K~\cite{ChengPAMI} &10,000&\\
\hline
\multirow{11}{*}{\rotatebox{90}{2018}}
&RADF~\cite{hu2018recurrently}  &AAAI &FCN &VGGNet
    &Fully-Sup. &STL &Object &MSRA10K~\cite{ChengPAMI} &10,000 &\checkmark\\
&ASMO~\cite{li2018weakly}  &AAAI &FCN &ResNet101
    &Weakly-Sup. &MTL &Object &MS COCO~\cite{lin2014microsoft}+MSRA-B~\cite{Liu2007}+HKU-IS~\cite{li2015visual} &82,783+2,500+2,500 &\checkmark\\
&LICNN~\cite{cao2018lateral}&AAAI &FCN &VGGNet
    &Weakly-Sup. &STL &Object &ImageNet~\cite{deng2009imagenet} &456k&\\
&BDMP~\cite{zhang2018bi} &CVPR &FCN &VGGNet
    &Fully-Sup. &STL &Object &DUTS~\cite{wang2017} &10,553&\\
&DUS~\cite{zhang2018deep}&CVPR &FCN &ResNet101
    &Un-Sup. &MTL &Object &MSRA-B~\cite{Liu2007} &2,500&\\
&DGRL~\cite{wang2018detect}&CVPR &FCN &ResNet50
    &Fully-Sup. &STL &Object &DUTS~\cite{wang2017} &10,553&\\
&PAGR~\cite{zhang2018progressive}&CVPR &FCN &VGGNet19
    &Fully-Sup. &STL &Object &DUTS~\cite{wang2017} &10,553&\\
&RSDNet~\cite{Islam_2018_CVPR}&CVPR &FCN &ResNet101
    &Fully-Sup. &MTL &Object &PASCAL-S~\cite{li2014secrets} &425&\\
&ASNet~\cite{Wang_2018_CVPR}&CVPR &FCN &VGGNet	
    &Fully-Sup. &MTL &Object &SALICON~\cite{jiang2015salicon}+MSRA10K~\cite{ChengPAMI}+DUT-OMRON~\cite{yang2013} &15,000+10,000+5,168&\\
&PiCANet~\cite{liu2018picanet}&CVPR&FCN &VGGNet/ResNet50
    &Fully-Sup. &STL &Object &DUTS~\cite{wang2017} &10,553 &\checkmark\\
&C2S-Net~\cite{li2018contour}&ECCV&FCN &VGGNet
    &Weakly-Sup. &MTL &Object &MSRA10K~\cite{ChengPAMI}+Web &10,000+20,000&\\
&RAS~\cite{chen2018reverse}&ECCV&FCN &VGGNet
    &Fully-Sup. &STL &Object &MSRA-B~\cite{Liu2007} &2,500&\\
\hline
\multirow{18}{*}{\rotatebox{90}{2019}}
&SuperVAE\cite{li2019supervae} &AAAI &FCN &N/A 
    &Un-Sup. &STL &Object &N/A &N/A&\\

&DEF\cite{zhuge2019deep} &AAAI &FCN &ResNet101
    &Fully-Sup. &STL &Object &DUTS~\cite{wang2017} &10,553&\\

&AFNet\cite{feng2019attentive} &CVPR &FCN &VGGNet16
    &Fully-Sup. &MTL &Object &DUTS~\cite{wang2017} &10,553&\\

&BASNet\cite{qin2019basnet} &CVPR &FCN &ResNet-34
    &Fully-Sup. &STL &Object &DUTS~\cite{wang2017} &10,553&\\

&CapSal\cite{zhang2019capsal} &CVPR &FCN &ResNet101
    &Fully-Sup. &MTL &Object &COCO-CapSal~\cite{zhang2019capsal}/DUTS~\cite{wang2017} &5,265/10,553 &\\

&CPD-R\cite{wu2019cascaded} &CVPR &FCN &ResNet50 
    &Fully-Sup. &STL &Object &DUTS~\cite{wang2017} &10,553&\\

&MLSLNet\cite{wu2019mutual} &CVPR &FCN &VGG16
    &Fully-Sup. &MTL &Object &DUTS~\cite{wang2017} &10,553&\\

&$^\dagger$MWS\cite{zeng2019multi} &CVPR &FCN &N/A
    &Weakly-Sup. &STL &Object &\tabincell{c}{ImageNet DET~\cite{deng2009imagenet}+MS COCO~\cite{lin2014microsoft}\\+ImageNet~\cite{Krizhevsky2012}+DUTS~\cite{wang2017}}
&\tabincell{c}{456k+82,783\\+300,000+10,553}&\\

&PAGE-Net\cite{wang2019salient} &CVPR &FCN  &VGGNet16
    &Fully-Sup. &MTL &Object &MSRA10K~\cite{ChengPAMI} &10,000&\checkmark\\

&PS\cite{wang2019iterative} &CVPR &FCN &ResNet50
    &Fully-Sup. &STL &Object &MSRA10K~\cite{ChengPAMI} &10,000&\checkmark\\

&PoolNet\cite{liu2019simple} &CVPR &FCN &ResNet50 
    &Fully-Sup. &STL/MTL &Object &DUTS~\cite{wang2017} &10,553&\\


&BANet\cite{su2019selectivity} &ICCV &FCN &ResNet50
    &Fully-Sup. &MTL &Object &DUTS~\cite{wang2017} &10,553&\\

&EGNet\cite{zhao2019egnet}   &ICCV &FCN &VGGNet/ResNet
    &Fully-Sup. &MTL &Object &DUTS~\cite{wang2017} &10,553&\\

&HRSOD\cite{zeng2019towards}  &ICCV &FCN &VGGNet
    &Fully-Sup. &STL &Object &DUTS~\cite{wang2017}/HRSOD~\cite{zeng2019towards}+DUTS~\cite{wang2017}
&10,553/12,163&\\

&JDFPR\cite{xu2019structured}   &ICCV &FCN &VGG
    &Fully-Sup. &STL &Object &MSRA-B~\cite{Liu2007} &2,500&\checkmark\\

&SCRN\cite{wu2019stacked}    &ICCV &FCN &ResNet50
    &Fully-Sup. &MTL &Object &DUTS~\cite{wang2017} &10,553&\\

&SSNet\cite{zeng2019joint}   &ICCV &FCN &Desenet169
    &Fully-Sup. &MTL &Object &PASCAL VOC 2012~\cite{everingham2010pascal}+DUTS~\cite{wang2017}
    &1,464+10,553&\checkmark\\ 
&TSPOANet\cite{liu2019employing}  &ICCV &Capsule &FLNet
    &Fully-Sup. &STL &Object &DUTS~\cite{wang2017} &10,553&\\
\hline
\end{tabular}
}
\end{threeparttable}
\end{table*}

The rest of the paper is organized as follows. \S\ref{sec:sod_models} explains the proposed taxonomies, {each accompanied with one or two most representative models.} \S\ref{sec:soddatasets} examines the most notable SOD datasets, whereas \S\ref{sec:evametric} describes several widely used SOD metrics. \S\ref{sec:benchmarking} benchmarks several deep SOD models and provides in-depth analyses. \S\ref{sec:discussions}
provides further discussions and presents open issues and future research directions of the field. Finally, \S\ref{sec:conclusion} concludes the paper.

\section{Deep Learning based SOD Models}
\label{sec:sod_models}
Before reviewing recent deep SOD models in details, we first {provide} a common formulation of the image-based SOD problem. Given an input image
$\bm{I}\!\in\!\mathbb{R}^{W\!\times \!H\!\times\!3\!}$ of size $W\!\times\!H$, an SOD model $f$ maps the input image $\bm{I}$ to a continuous saliency map $\bm{S}\!=\!f(\bm{I})\!\in\![0,1]^{W\!\times \!H\!}$.
For learning-based SOD, the model $f$ is learned through a set of training samples. Given a set of static images $\mathcal{I}\!\!=\!\!\{\bm{I}_{n\!}\!\in\!\mathbb{R}^{W\!\times \!H\!\times\!3}\}_{n}$ and corresponding binary SOD ground-truth masks $\mathcal{G}\!\!=\!\!\{\bm{G}_{n\!}\!\in\!\{0,1\}^{W\!\times\! H}\}_{n}$, the goal of learning is to find $f\!\in\!\mathcal{F}$ that minimizes the prediction error, \ie, $\sum_{n\!}\ell(\bm{S}_n, \bm{G}_n)$, where $\ell$ is a certain distance measure (\eg, defined in \S\ref{sec:evametric}), $\bm{S}_{n\!}\!=_{\!}\!f(\bm{I}_n)$, and $\mathcal{F}$ is the set of potential mapping functions. Deep SOD methods typically model $f$ through modern deep learning techniques, as will be reviewed later in this section. The ground-truths $\mathcal{G}$ can be collected by different methodologies, \ie, direct human-annotation or eye-fixation-guided labeling, and may have different formats, \ie, pixel-wise or bounding-box annotations, which will be discussed in \S\ref{sec:soddatasets}.

{In Table~\ref{table:sod_texomonies}, we categorize recent deep SOD models according to four taxonomies, considering \textit{network architecture} (\S\ref{sec:sod_network_structure}), \textit{level of supervision} (\S\ref{sec:sod_supervision}), \textit{learning paradigm} (\S\ref{sec:sod_learning_paradigm}), and {whether they works at an \textit{object or instance level}} (\S\ref{sec:sod_pixel_instance}). In the following, each category is elaborated {on} and exemplified by one or two most representative models. Table~\ref{table:sod_methods} summarizes essential characteristics of recent SOD models.}

\subsection{Representative Network Architectures for SOD}\label{sec:sod_network_structure}
Based on the primary network architectures adopted, we classify deep SOD models into four categories, namely \textit{MLP}-based (\S\ref{sec:mlp}), \textit{FCN}-based (\S\ref{sec:fcn}), \textit{hybrid network}-based (\S\ref{sec:hybrid}) and \textit{Capsule}-based (\S\ref{sec:capsule}).

\begin{figure*}[t]
  \centering
      \includegraphics[width=0.9 \linewidth]{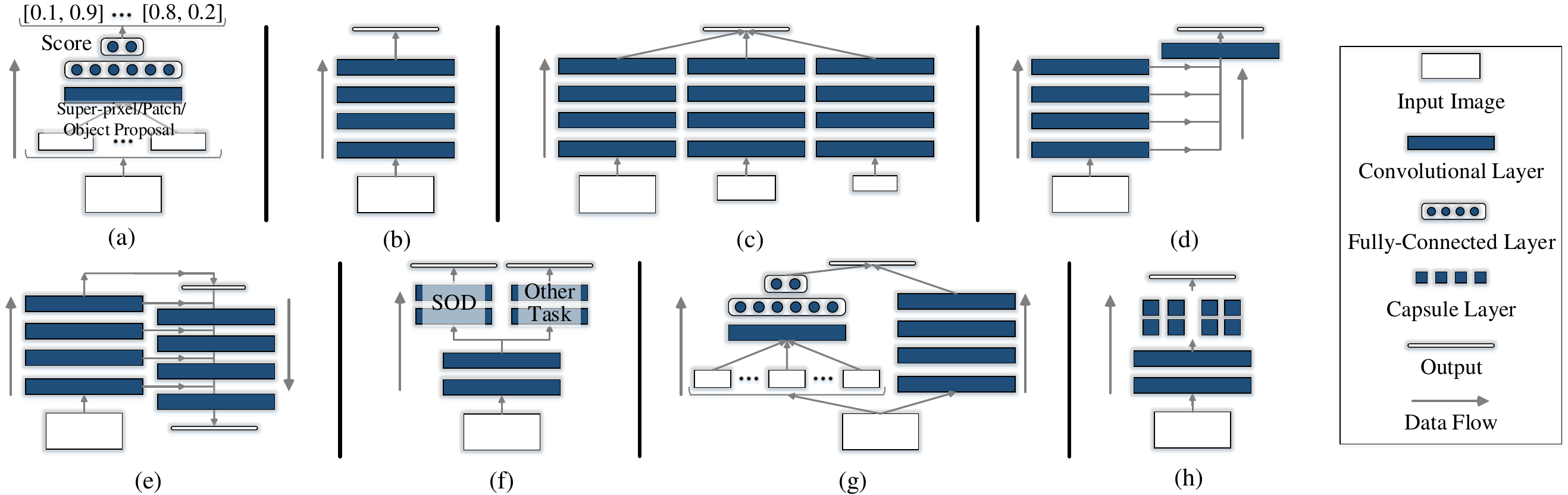}
\caption{{Categorization} of previous deep SOD models according to the adopted network architecture. (a) MLP-based methods. (b)-(f) FCN-based methods, mainly using (b) single-stream network, (c) multi-stream network, (d) side-out fusion network, (e)  bottom-up/top-down network, and (f) branch network architectures.  (g) Hybrid network-based methods. (h) Capsule-based methods.  See \S\ref{sec:sod_network_structure} for more detailed descriptions.}
\label{fig:SODcategory}
\end{figure*}

\subsubsection{Multi-Layer Perceptron (MLP)-Based Methods}\label{sec:mlp}

MLP-based methods leverage image subunits (\ie, \emph{super-pixels/patches}~\cite{zhao2015saliency,he2015supercnn,lee2016deep} and generic \emph{object proposals}~\cite{wang2015deep,li2015visual,zhang2016unconstrained,kim2016shape}) as processing units. They feed deep features extracted from the subunits into an MLP-classifier for saliency score prediction (Fig.~\ref{fig:SODcategory}(a)).

\vspace{1.mm}\noindent\textbf{1) Super-pixel/patch-based methods} use regular (patch) or nearly-regular (super-pixel) image decomposition.
As an example of regular decomposition, MCDL~\cite{zhao2015saliency} uses two pathways {to extract} local and global context from two super-pixel-centered windows of different sizes. The global and local feature vectors are fed into an MLP for classifying background and saliency.
In contrast, SuperCNN~\cite{he2015supercnn} constructs two hand-crafted input feature sequences for each irregular super-pixel, and use two separate CNN columns to produce saliency scores from the feature sequences, respectively.
Regular image decomposition can accelerate the processing speed, thus most of the methods in this category are based on regular decompostion.

\vspace{1.mm}\noindent\textbf{2) Object proposal-based methods} leverage object proposals~\cite{wang2015deep,li2015visual} or 
bounding-boxes~\cite{zhang2016unconstrained,kim2016shape} as basic processing units in order to better encode object information.
For instance, MAP~\cite{zhang2016unconstrained} uses a CNN model to generate a set of scored bounding-boxes, then selects an optimized compact subset of bounding-boxes as the salient objects.
Note that this kind of methods typically produce coarse SOD results due to the lack of object boundary information.

Though MLP-based SOD methods {greatly} outperform {their} non-deep counterparts, they  cannot fully leverage essential spatial information and are quite time-consuming, as they need to process all visual subunits one-by-one.

\subsubsection{Fully Convolutional Network (FCN)-Based Methods}\label{sec:fcn}
To address the limitations of MLP-based methods, recent solutions adopt FCN architecture~\cite{long2015fully}, leading to end-to-end spatial saliency representation learning and fast saliency prediction, {within a single feed-forward process. FCN-based methods are now dominant in the field.} {Typical architectures can be further classified as: \emph{single-stream}, \emph{multi-stream}, \emph{side-fusion}, \emph{bottom-up/top-down}, and \emph{branched networks}.}

\vspace{1.mm}\noindent\textbf{1) Single-stream network} is the most standard architecture, {having} a stack of convolutional layers, {interleaved} with pooling and non-linear activation operations (see Fig.~\ref{fig:SODcategory}(b)).
{It takes a whole image as input, and directly outputs a pixel-wise probabilistic map highlighting salient objects.} For example, UCF~\cite{Zhang_2017_ICCV} makes use of an encoder-decoder network architecture for finer-resolution saliency prediction. It incorporates a reformulated dropout in the encoder {to learn} uncertain features, and a hybrid {upsampling} scheme in the decoder {to avoid} checkerboard artifacts.

\vspace{1.mm}\noindent\textbf{2) Multi-stream network}, as depicted in Fig.~\ref{fig:SODcategory}(c), typically consists of multiple network streams to explicitly {learn} multi-scale saliency features from multi-resolution inputs. Multi-stream outputs are fused to form a final prediction.  {DCL~\cite{li2016deep}, as one of the earliest attempts towards this direction}, contains two streams, which produce pixel- and region-level SOD estimations, respectively.

\vspace{1.mm}\noindent\textbf{3) Side-fusion network} fuses multi-layer responses of a backbone network together for SOD prediction, making use of the complementary information of the inherent multi-scale representations of the CNN hierarchy (Fig.~\ref{fig:SODcategory}(d)). Side-outputs are typically supervised by the ground-truth, leading to a \textit{deep supervision} strategy~\cite{xie2015holistically}. {As a well-known side-fusion network based SOD model,} DSS~\cite{hou2016deeply}$_{\!}$ adds$_{\!}$ short connections from deeper side-outputs to shallower ones. In this way, higher-level features help lower side-outputs {to} better locate salient regions, and lower-level$_{\!}$ features$_{\!}$ can$_{\!}$ enrich$_{\!}$ deeper$_{\!}$ side-outputs$_{\!}$ with$_{\!}$ finer$_{\!}$ details.

\vspace{1.mm}\noindent\textbf{4) Bottom-up/top-down network} refines rough saliency maps in the feed-forward pass by gradually incorporating spatial-detail-rich features from lower layers, and produces the finest saliency maps at the top-most layer (Fig.~\ref{fig:SODcategory}(e)), which resembles the U-Net~\cite{ronneberger2015u} for semantic segmentation. {This network architectures is first adopted by
PiCANet~\cite{liu2018picanet}, which hierarchically embeds global and local pixel-wise attention modules to selectively attend to informative context.}

\vspace{1.mm}\noindent\textbf{5) Branched network} typically addresses multi-task learning for more robust saliency pattern modeling. They have a \textit{single-input-multiple-output}
structure, where bottom layers are shared to process a common input and top ones are specialized for different tasks (Fig.~\ref{fig:SODcategory}(f)). For example, C2S-Net~\cite{li2018contour} is constructed by adding a pre-trained contour detection model~\cite{yang2016object} to a main SOD branch. Then the two branches are alternately trained for the two tasks, \ie, SOD and contour detection.

\subsubsection{Hybrid Network-Based Methods}\label{sec:hybrid}
Some other models combine both MLP- and FCN-based subnets to produce edge-preserving results with multi-scale context (Fig.~\ref{fig:SODcategory}(g)).
{{Combining} pixel-level and region-level saliency cues is {a} promising {strategy} to yield improved performance, {though it introduces} extra computational costs.}
{CRPSD~\cite{tang2016saliency} consolidates this idea. It combines pixel- and region-level saliency.} The former is generated by fusing the last and penultimate side-output features of an FCN, while the latter is obtained by applying an existing SOD model~\cite{zhao2015saliency} to image regions. Only the FCN and fusion layers are trainable.

\subsubsection{Capsule-Based Methods}\label{sec:capsule}
Recently, Hinton~\etal~\cite{hinton2011transforming} proposed a new {family of neural networks}, named \emph{Capsules}. Capsules are made up of a group of neurons which accept and output vectors as opposed to scalar values of CNNs, allowing {entity properties to be comprehensively modeled}. Some researchers have thus been inspired to explore Capsules in SOD~\cite{qi2019multi,liu2019employing} (Fig.~\ref{fig:SODcategory}(h)). For instance, TSPOANet~\cite{liu2019employing} emphasizes part-object relations {using} a two-stream capsule network. The input features of capsules are extracted from a CNN, {and} transformed into low-level capsules. {These are then} assigned to high-level {capsules}, and finally recognized to be salient or background.

\subsection{Level of Supervision}\label{sec:sod_supervision}
Based on the {type} of supervision, deep SOD models can be classified into either \emph{fully-supervised} or {\emph{weakly-/unsupervised}}.

\subsubsection{Fully-Supervised Methods}
Most deep SOD models are trained with large-scale {pixel-level} human annotations, {which are} time-consuming and expensive {to acquire}. Moreover, models trained on fine-labeled datasets tend to overfit and generalize poorly to real-life images~\cite{zhang2018deep}.
Thus, training SOD with weaker annotations {has become} an increasingly popular research direction.

\subsubsection{{Weakly-/Unsupervised Methods}}
To get rid of laborious manual labeling, {several weak supervision forms have been explored in SOD}, including \emph{image-level category} labels~\cite{wang2017,cao2018lateral}, \emph{object contours}~\cite{li2018contour}, \emph{image captions}~\cite{zeng2019multi} and \emph{pseudo ground-truth} masks generated by non-learning SOD methods~\cite{zhang2017supervision,zhang2018deep,li2018weakly}.

\vspace{1.mm}\noindent\textbf{1) Category-level supervision.} It {has been shown} that deep features trained with only image-level labels {also provide information on} object locations~\cite{long2014convnets,zhou2016learning}, {making them} promising supervision signals for SOD training. {WSS~\cite{wang2017}, as a typical example, first pre-trains a two-branch network,} {where one branch is used to predict image labels based on} ImageNet~\cite{deng2009imagenet}, {and the other estimates} SOD maps. The estimated maps are refined by CRF and used to further fine-tune the SOD branch.

\vspace{1.mm}\noindent\textbf{2) Pseudo pixel-level supervision.}
Though informative, image-level labels are weak. Some researchers {therefore} instead use traditional non-learning SOD methods~\cite{zhang2017supervision,zhang2018deep,li2018weakly}, or contour information~\cite{li2018contour}, to generate noisy yet finer-grained cues for training.
For instance, SBF~\cite{zhang2017supervision} fuses weak saliency maps from a set of prior heuristic SOD models~\cite{yan2013hierarchical,zhang2015minimum,zhang2016exploiting} at intra- and inter-image levels, to generate supervision signals. C2S-Net~\cite{li2018contour} trains the SOD branch with the pixel-wise salient object masks generated from the outputs of the contour branch~\cite{arbelaez2014multiscale} using CEDN~\cite{yang2016object}. The contour and SOD branches alternatively update each other and progressively output finer SOD predictions.

\subsection{Learning Paradigm}\label{sec:sod_learning_paradigm}

From the perspective of learning paradigms, SOD networks can be divided into \emph{single-task learning (STL)} and \emph{multi-task learning (MTL)} methods.

\subsubsection{Single-Task Learning (STL) Based Methods}

In machine learning, the standard {practice} is to learn one task at a time~\cite{caruana1997multitask}, \ie, STL. Most deep SOD methods belong to this realm of learning, \ie, {they} utilize supervision from a single knowledge domain (SOD or anther related field such as image classification~\cite{cao2018lateral}) for training.

\subsubsection{Multi-Task Learning (MTL) Based Methods}
Inspired by the human learning process, where knowledge learned from related tasks can assist the learning of a new task, MTL~\cite{caruana1997multitask} aims to improve the performance of multiple related tasks by learning them simultaneously.
Benefiting from extra knowledge from related tasks, models can gain improved generalizability. An extra advantage lies in the sharing of samples among tasks, which alleviates the lack of data for training heavily parameterized models. {These are the core motivations of MTL based SOD models, and branched architectures (see \S\ref{sec:fcn}) are usually adopted.}

\vspace{1.mm}\noindent\textbf{1) Salient object subitizing}.
The ability of humans to rapidly enumerate a small number of items is known as subitizing~\cite{kaufman1949discrimination,zhang2015salient}. {Inspired by this}, some works learn salient object subitizing and detection simultaneously~\cite{zhang2016unconstrained,he2017delving,Islam_2018_CVPR}.
{{RSDNet~\cite{Islam_2018_CVPR}} represents the latest advance in this direction. It addresses detection, ranking and subitizing of salient objects in a unified framework.}

\vspace{1.mm}\noindent\textbf{2) Fixation prediction}
aims to predict human eye-fixation locations in visual scenes. Due to its close relation with SOD, learning shared knowledge from these two tasks can improve the performance of both. For example, ASNet~\cite{Wang_2018_CVPR} derives fixation information as a high-level understanding of the scene, from upper network layers. Then, fine-grained object-level saliency is progressively optimized {under} the guidance of the fixation in a top-down manner.
\vspace{1.mm}\noindent\textbf{3) Image classification}.
Image-level tags are valuable for SOD, as they provide the category
information of dominant objects in the images which are
very likely to be the salient regions~\cite{wang2017}. Inspired by this, some SOD models learn image classification as an auxiliary task.
For example, ASMO~\cite{li2018weakly} leverages class activation maps from a neural classifier and saliency maps from {previous} non-learning SOD methods to train the SOD network, in an iterative manner.

\vspace{1.mm}\noindent\textbf{4) Semantic segmentation} is for per-pixel semantic prediction.  Though SOD is class-agnostic, high-level semantics play a crucial role in saliency modeling. Thus, the task of semantic segmentation can also be integrated into SOD learning.
{A recent SOD model, SSNet~\cite{zeng2019joint}, is developed upon this idea.} It  uses a saliency aggregation module to predict a saliency score of each category. Then, a segmentation network is used to produce segmentation masks of all the categories. These masks are finally aggregated (according to corresponding saliency scores) to produce a SOD map.

\begin{table*}
\centering
\caption{Statistics of popular SOD datasets,
        {including the number of images, number of salient objects per image, area ratio of the salient objects {in} images, annotation type, image resolution, and existence of fixation data. }
        See \S\ref{sec:soddatasets} for more detailed descriptions.}
\begin{threeparttable}
\resizebox{1\textwidth}{!}{
\setlength\tabcolsep{2pt}
\renewcommand\arraystretch{1.05}
\begin{tabular}{|c|r||c|c|c|c|c|c|ll|c|}
\hline\thickhline
&Dataset~~~~~~~~~&Year &Publ. &\#Img. &\#Obj. &Obj. Area(\%) &SOD Annotation &\multicolumn{2}{c|}{Resolution} &Fix.\\
\hline
\hline
\multirow{5}{*}{\rotatebox{90}{Early}}
&MSRA-A~\cite{Liu2007} &2007 &CVPR  &1,000/20,840 &1-2 &- &bounding-box object-level &~~~~~~~~~~~~~~~~~~~~~~~~~~- 
&&\\
&MSRA-B~\cite{Liu2007} &2007 &CVPR &5,000 &1-2 &20.82$_{\pm 10.29}$  &bounding-box object-level, pixel-wise object-level &$\max(w,h)\!=\!400, $&$\min(w,h)\!=\!126$  &\\
&SED1~\cite{alpert2007image} &2007 &CVPR &100 &1 &26.70$_{\pm 14.26}$  &pixel-wise object-level &$\max(w,h)\!=\!465, $&$\min(w,h)\!=\!125$  & \\
&SED2~\cite{alpert2007image} &2007 &CVPR &100 &2 &21.42$_{\pm 18.41}$  &pixel-wise object-level &$\max(w,h)\!=\!300, $&$\min(w,h)\!=\!144$  & \\
&ASD~\cite{achanta2009frequency} &2009 &CVPR &1,000 &1-2 &19.89$_{\pm 9.53}$  &pixel-wise object-level &$\max(w,h)\!=\!400, $&$\min(w,h)\!=\!142$ & \\
\hline
\multirow{7}{*}{\rotatebox{90}{Modern\&Popular}}
&SOD~\cite{Movahedi2010} &2010 &CVPR-W &300 &1-4+ &27.99$_{\pm 19.36}$ &pixel-wise object-level &$\max(w,h)\!=\!481, $&$\min(w,h)\!=\!321$ & \\
&MSRA10K~\cite{ChengPAMI}&2015 &TPAMI  &10,000 &1-2 &22.21$_{\pm 10.09}$ &pixel-wise object-level &$\max(w,h)\!=\!400, $&$\min(w,h)\!=\!144$ & \\
&ECSSD~\cite{shi2015hierarchical} &2015 &TPAMI  &1,000 &1-4+ &23.51$_{\pm 14.02}$ &pixel-wise object-level &$\max(w,h)\!=\!400, $&$\min(w,h)\!=\!139$ &\\
&DUT-OMRON~\cite{yang2013} &2013 &CVPR  &5,168 &1-4+ &14.85$_{\pm 12.15}$ &pixel-wise object-level &$\max(w,h)\!=\!401, $&$\min(w,h)\!=\!139$ &\checkmark\\
&PASCAL-S~\cite{li2014secrets}  &2014 &CVPR    &850  &1-4+ &24.23$_{\pm 16.70}$ &pixel-wise object-level &$\max(w,h)\!=\!500, $&$\min(w,h)\!=\!139$ &\checkmark\\
&HKU-IS~\cite{li2015visual}   &2015  &CVPR    &4,447 &1-4+ &19.13$_{\pm 10.90}$ &pixel-wise object-level &$\max(w,h)\!=\!500, $&$\min(w,h)\!=\!100$ &\\
&DUTS~\cite{wang2017} &2017 &CVPR &15,572 &1-4+ &23.17$_{\pm 15.52}$ &pixel-wise object-level &$\max(w,h)\!=\!500,$&$\min(w,h)\!=\!100$ &\\
\hline
\multirow{7}{*}{\rotatebox{90}{Special}}
&SOS~\cite{zhang2015salient} &2015 &CVPR &6,900 &0-4+ &41.22$_{\pm 25.35}$ &object number, bounding-box (\textit{train} set) &$\max(w,h)\!=\!6132,$&$\min(w,h)\!=\!80~~$  &\\
&MSO~\cite{zhang2015salient}&2015  &CVPR &1,224 &0-4+ &39.51$_{\pm 24.85}$ &object number, bounding-box instance-level&$\max(w,h)\!=\!3888,$&$\min(w,h)\!=\!120$  &  \\
&ILSO~\cite{li2017instance} &2017 &CVPR &1,000 &1-4+ &24.89$_{\pm 12.59}$ &pixel-wise instance-level &$\max(w,h)\!=\!400, $&$\min(w,h)\!=\!142$  & \\ 
&XPIE~\cite{xia2017and} &2017 &CVPR &10,000 &1-4+ &19.42$_{\pm 14.39}$ &pixel-wise object-level,  geographic information &$\max(w,h)\!=\!500, $&$\min(w,h)\!=\!130$  &\checkmark\\
&SOC~\cite{Fan_2018_ECCV} &2018 &ECCV &6,000 &0-4+ &21.36$_{\pm 16.88}$ &pixel-wise instance-level, object category, attribute &$\max(w,h)\!=\!849, $&$\min(w,h)\!=\!161$  & \\ 
&COCO-CapSal~\cite{zhang2019capsal}&2019 &CVPR &6,724 &1-4+ &23.74$_{\pm 17.00}$ &pixel-wise object-level, image caption &$\max(w,h)\!=\!640, $&$\min(w,h)\!=\!480$ &\\
&HRSOD~\cite{zeng2019towards} &2019 &ICCV &2,010 &1-4+ &21.13$_{\pm 15.14}$ &pixel-wise object-level &$\max(w,h)\!=\!10240, $&$\min(w,h)\!=\!600$ & \\
\hline
\end{tabular}
}
\end{threeparttable}
\label{table:dataset}
\end{table*}

\vspace{1.mm}\noindent\textbf{5) Contour/edge detection}
refers to the task of detecting obvious object boundaries in images, which are informative of salient objects. Thus, it is also explored in SOD modeling. For example, PAGE-Net~\cite{wang2019salient} learns an edge detection module and embeds edge cues into the main SOD stream in a top-down manner, leading to better edge-preserving results.

\vspace{1.mm}\noindent\textbf{6) Image Captioning} can provide extra knowledge about the main content of visual scenes, enabling SOD models to better capture high-level semantics. {This has been explored in
CapSal~\cite{zhang2019capsal}, which incorporates semantic context from a captioning network with local-global visual cues to achieve improved performance for detecting salient objects.}

\subsection{Object-/Instance-Level SOD}\label{sec:sod_pixel_instance}
According to whether {or not they can} identify different salient object instances,
current deep SOD models can be categorized into \textit{object-level} and \textit{instance-level} methods.

\subsubsection{Object-Level Methods}
Most deep SOD models are object-level methods, \ie, designed to detect pixels that belong to salient objects without being aware of individual object instances.

\subsubsection{Instance-Level Methods}
Instance-level SOD methods further identify individual object instances in the detected salient regions, which is crucial for practical applications that need finer distinctions, such as semantic segmentation~\cite{fan2018associating} and multi-human parsing~\cite{zhao2019fine}. {As an early attempt, MSRNet~\cite{li2017instance} performs salient instance detection by decomposing it into three sub-tasks,} \ie, pixel-level saliency prediction, salient object contour detection and salient instance identification. It jointly performs the first two sub-tasks by integrating deep features for several different scaled versions of the input image. The last sub-task is solved by multi-scale combinatorial grouping~\cite{arbelaez2014multiscale} to generate salient object proposals from the detected contours and filter out noisy or overlapping ones.

\section{SOD Datasets}\label{sec:soddatasets}
With the rapid development of SOD, numerous datasets have been introduced.
Table~\ref{table:dataset} summarizes $19$ SOD datasets, which are highly representative  and widely used for training or benchmarking, or collected with specific properties.

\subsection{{Quick Overview}}\label{sec:sod_bmk_quick}
{In an attempt to facilitate understanding of SOD datasets, we present some main take-away points of this section. }

\noindent{\textbullet~Compared with early datasets~\cite{Liu2007,alpert2007image,achanta2009frequency}, recent~ones \cite{ChengPAMI,yang2013,li2015visual,wang2017}
are typically more advanced with less center bias, improved complexity, and increased scale. {They are thus better-suited} for training and evaluation,  and likely to have longer life-spans.}

\noindent{\textbullet~Some other recent datasets~\cite{zhang2015salient,li2017instance,xia2017and,Fan_2018_ECCV,zhang2019capsal,zeng2019towards}
are enriched with more diverse annotations (\eg, subitizing, captioning), representing new trends in the field.}

More$_{\!}$ in-depth$_{\!}$ discussions$_{\!}$ regarding$_{\!}$ generalizability$_{\!}$ and$_{\!}$ {difficulty}$_{\!}$ of$_{\!}$ several famous$_{\!}$ datasets will$_{\!}$ be$_{\!}$ presented$_{\!}$ in$_{\!}$  \S\ref{sec:cd_generalization}.\!\!

\subsection{Early SOD Datasets}\label{sec:early_sod_bmk}
Early SOD datasets typically contain simple scenes where 1-2 salient objects stand out from a clear background.

\noindent\textbullet~\textbf{MSRA-A~\cite{Liu2007}} contains 20,840 images. Each image has only one noticeable and eye-{catching} object, annotated by a bounding-box. As a subset of MSRA-A, MSRA-B has 5,000 images and less ambiguity \wrt the salient object.

\noindent\textbullet~\textbf{SED~\cite{alpert2007image}}\footnote{\myfont{\url{http://www.wisdom.weizmann.ac.il/~vision/Seg_Evaluation_DB}}} comprises a single-object subset and a two-object subset; each has 100 images with mask annotations.

\noindent\textbullet~\textbf{ASD~\cite{achanta2009frequency}}\footnote{\url{https://ivrlwww.epfl.ch/supplementary_material/RK_CVPR09/}}, also a subset of MSRA-A, has 1,000 images with pixel-wise ground-truths.

\subsection{Popular Modern SOD Datasets}\label{sec:modern_sod_bmk}
Recent SOD datasets tend to include more challenging and general scenes with relatively complex backgrounds and$_{\!}$ multiple$_{\!}$ salient$_{\!}$ objects. All$_{\!}$ have$_{\!}$ pixel-wise$_{\!}$ annotations.

\noindent\textbullet~\textbf{SOD~\cite{Movahedi2010}}\footnote{\url{http://elderlab.yorku.ca/SOD/}} consists of 300 images, constructed from~\cite{martin2001database}. Many images have more than one salient object that is similar to the background or touches image boundaries.

\noindent\textbullet~\textbf{MSRA10K~\cite{ChengPAMI}}\footnote{\url{https://mmcheng.net/zh/msra10k/}}, also known as THUS10K, contains 10,000 images selected from MSRA-A and covers all the images in ASD. Due to its large scale, MSRA10K is widely used to train deep SOD models (see Table~\ref{table:sod_methods}).

\noindent\textbullet~\textbf{ECSSD~\cite{shi2015hierarchical}}\footnote{\url{http://www.cse.cuhk.edu.hk/leojia/projects/hsaliency}}
is composed of 1,000 images with semantically
meaningful but structurally complex natural contents.

\noindent\textbullet~\textbf{DUT-OMRON~\cite{yang2013}}\footnote{\url{http://saliencydetection.net/dut-omron/}} has 5,168 images of complex backgrounds and diverse content, with pixel-wise annotations.

\noindent\textbullet~\textbf{PASCAL-S~\cite{li2014secrets}}\footnote{\url{http://cbi.gatech.edu/salobj/}\!} {comprises} 850 challenging images selected from the PASCAL VOC2010 \textit{val} set~\cite{everingham2010pascal}. With$_{\!}$ eye-fixation$_{\!}$ records, non-binary$_{\!}$ salient-object$_{\!}$ mask$_{\!}$ annotations$_{\!}$ are provided. {Note that}$_{\!}$ the$_{\!}$ saliency$_{\!}$ value$_{\!}$ of$_{\!}$ a$_{\!}$ pixel$_{\!}$ is$_{\!}$ calculated$_{\!}$ as$_{\!}$ the$_{\!}$ ratio$_{\!}$ of$_{\!}$ subjects$_{\!}$ that$_{\!}$ select$_{\!}$ the$_{\!}$ segment$_{\!}$ containing$_{\!}$ this$_{\!}$ pixel$_{\!}$ as$_{\!}$ salient$_{\!}$.

\noindent\textbullet~\textbf{HKU-IS~\cite{li2015visual}}\footnote{\url{https://i.cs.hku.hk/~gbli/deep_saliency.html}} has $4,447$ complex scenes that typically contain multiple disconnected objects with diverse spatial distributions and similar fore-/background appearances.

\noindent\textbullet~\textbf{DUTS~\cite{wang2017}}\footnote{\url{http://saliencydetection.net/duts/}} is a large-scale dataset, where the $10,553$  training~images were selected from the ImageNet \textit{train/val} set~\cite{deng2009imagenet}, and the $5,019$ test images are from the ImageNet \textit{test} set and SUN~\cite{xiao2010sun}. Since 2017, SOD models are typically trained on DUTS (Table~\ref{table:sod_methods}).

\subsection{Other Special SOD Datasets}\label{sec:special_sod_bmk}
{In addition to the}$_{\!}$ above ``standard" SOD datasets, some special ones {have also recently been proposed}, leading to new research directions.

\noindent\textbullet~\textbf{SOS~\cite{zhang2015salient}}\footnote{\url{http://cs-people.bu.edu/jmzhang/sos.html}} is created for SOD subitizing~\cite{kaufman1949discrimination}. It contains 6,900 images (\textit{training} set: 5,520, \textit{test} set: 1,380).
Each image is labeled as containing 0, 1, 2, 3 or 4+ salient objects.

\noindent\textbullet~\textbf{MSO~\cite{zhang2015salient}}\footnote{\url{http://cs-people.bu.edu/jmzhang/sos.html}} is a subset of SOS-\textit{test}~\cite{zhang2015salient}, covering 1,224 images. It has a more balanced distribution of the number of salient objects. Each object has a bounding-box annotation.

\noindent\textbullet~\textbf{ILSO~\cite{li2017instance}}\footnote{\url{http://www.sysu-hcp.net/instance-level-salient-object-segmentation/}}
contains 1,000 images with precise instance-level annotations and coarse contour labeling.

\noindent\textbullet~\textbf{XPIE~\cite{xia2017and}}\footnote{\url{http://cvteam.net/projects/CVPR17-ELE/ELE.html}\!}
has 10,000 images with pixel-wise labels. It has three subsets: \textit{Set-P} has 625 images of places-of-interest with geographic information; \textit{Set-I} 8,799 images with object tags; and \textit{Set-E} 576 images with eye-fixation records.

\noindent\textbullet~\textbf{SOC~\cite{Fan_2018_ECCV}}\footnote{\url{http://mmcheng.net/SOCBenchmark/}}
consists of 6,000 images with 80 common categories. Half of the images contain salient objects, {while the remaining} have none. Each {image containing salient objects} is annotated with an instance-level ground-truth mask, object category, and challenging factors. The non-salient object subset has 783 texture images and 2,217 real-scene images.

\noindent\textbullet~\textbf{COCO-CapSal~\cite{zhang2019capsal}}\footnote{\url{https://github.com/yi94code/HRSOD}}
is built from COCO~\cite{lin2014microsoft} and SALICON~\cite{jiang2015salicon}. Salient objects were first roughly localized using the mouse-click data in SALICON, then precisely annotated according to the instance masks in COCO. The dataset has 5,265 and 1,459 images for training and testing, respectively.

\noindent\textbullet~\textbf{HRSOD~\cite{zeng2019towards}}\footnote{\url{https://github.com/zhangludl/code-and-dataset-for-CapSal}}
is the first \emph{high-resolution} dataset for SOD. It contains 1,610 training and 400 testing images collected from websites. Pixel-wise ground-truths are provided.

\begin{figure}[t]
  \centering
      \includegraphics[width=1 \linewidth]{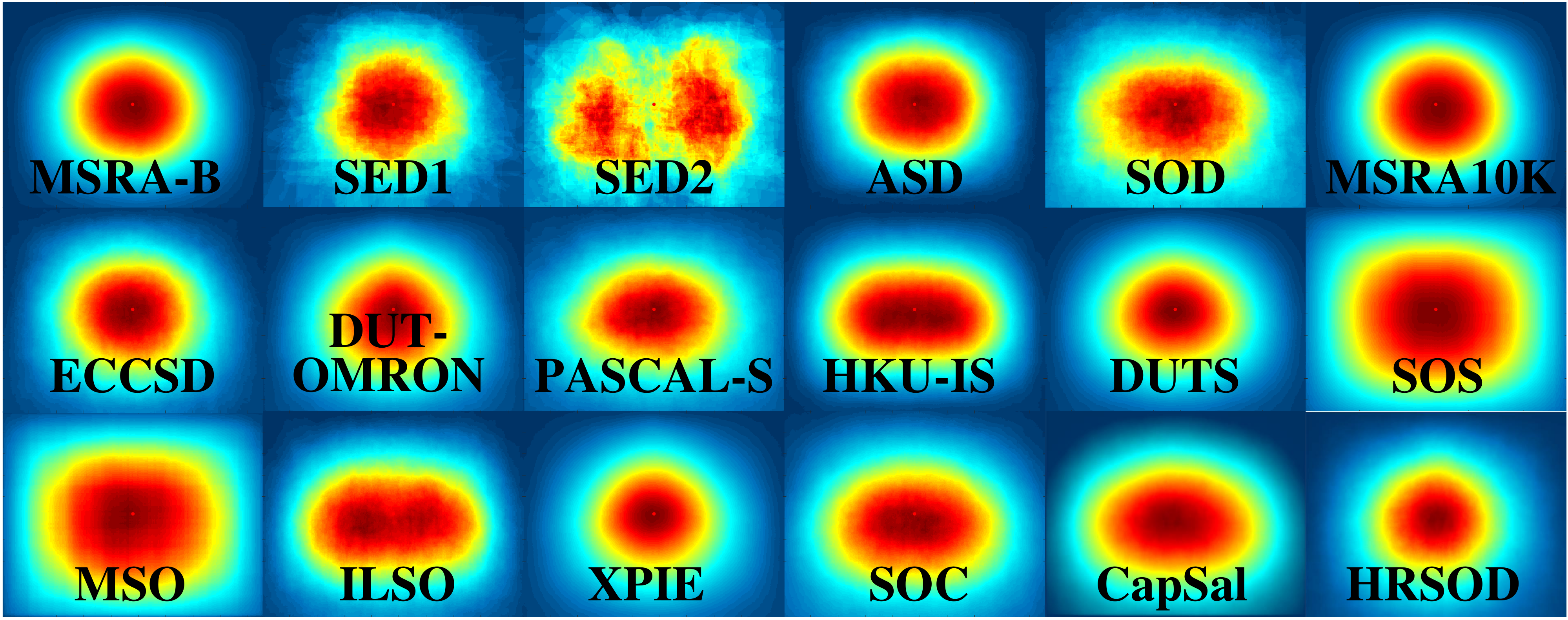}
\caption{Annotation distributions of SOD datasets (see \S\ref{sec:soddatasets} for details).}
\label{fig:centerbias}
\end{figure}

\subsection{{Discussion}}\label{sec:dataset_discussion}
As shown in Table~\ref{table:dataset}, \textit{early SOD datasets}~\cite{Liu2007,alpert2007image,achanta2009frequency} are comprised of simple images with 1-2 salient objects per image, and only provide rough bounding-box annotations, which are insufficient for reliable evaluations~\cite{achanta2009frequency,wang2008two}.
Performance on these datasets has become saturated. \textit{Modern datasets}~\cite{ChengPAMI,shi2015hierarchical,yang2013,li2015visual,wang2017} are typically large-scale and offer precise pixel-wise ground-truths. The scenes are more complex and general, and usually contain multiple salient objects. Some \textit{special datasets}~contain challenging scenes with background only \cite{zhang2015salient,Fan_2018_ECCV}, provide  more fine-grained, instance-level SOD ground-truths  \cite{li2017instance,Fan_2018_ECCV} or include other annotations such as image captions~\cite{zhang2019capsal}, inspiring new research directions and applications. Fig.~\ref{fig:centerbias} depicts the annotation distributions of $18$ SOD datasets. Here are some essential conclusions: 1) Some datasets~\cite{Liu2007,achanta2009frequency,ChengPAMI,wang2017} have significant center bias; 2) {Datasets} \cite{li2015visual,li2017instance,zhang2019capsal} have more balanced location distributions for salient objects; and 3)  MSO~\cite{zhang2015salient} has less center bias, as only bounding-box annotations are provided. We analyze the generalizability and difficulty of several famous SOD datasets in-depth in \S\ref{sec:cd_generalization}.

\section{Evaluation Metrics}\label{sec:evametric}
This section reviews popular object-level SOD evaluation metrics, \ie, Precision-Recall (PR), F-measure~\cite{achanta2009frequency}, Mean Absolute Error (MAE)~\cite{perazzi2012saliency}, weighted $F_{\beta}$ measure (Fbw)~\cite{margolin2014evaluate}, Structural measure (S-measure)~\cite{FanStructMeasureICCV17}, and Enhanced-alignment measure (E-measure)~\cite{Fan2018Enhanced}.

\subsection{{Quick Overview}}
{To better understand the characteristics of different metrics, a quick overview of the main conclusions for this section {are provided as follows}.}

\noindent{\textbullet~PR, F-measure, MAE, and Fbw address \textit{pixel-wise} errors, while S-measure and E-measure consider \textit{structure} cues.} 

\noindent{\textbullet~Among pixel-level metrics, PR, F-measure, and Fbw  fail to consider true negative pixels, while MAE can remedy this. }

\noindent{\textbullet~Among$_{\!}$ structured$_{\!}$ metrics, S-measure$_{\!}$ is$_{\!}$ more$_{\!}$ favored$_{\!}$ than$_{\!}$ E-measure,$_{\!}$  as$_{\!}$ SOD$_{\!}$ addresses$_{\!}$ continuous$_{\!}$ saliency$_{\!}$ estimates.}

\noindent{\textbullet~Considering popularity, advantages and completeness, F-measure, S-measure and MAE are the most recommended and$_{\!}$ are$_{\!}$ thus$_{\!}$ used$_{\!}$ for$_{\!}$ our$_{\!}$ performance$_{\!}$ benchmarking$_{\!}$ in$_{\!}$ \S\ref{sec:overall_bmk}.}

\subsection{{Metric Details}}
\noindent\textbullet~\textbf{PR} is calculated based on {the binarized} salient object mask and ground-truth:
\begin{equation}\label{equation:pr}\small
  \text{Precision}= \frac{\text{TP}}{\text{TP}+\text{FP}},~~~~\text{Recall}= \frac{\text{TP}}{\text{TP}+\text{FN}},
\end{equation}
where TP, TN, FP, FN denote true-positive, true-negative, false-positive, and false-negative, respectively.
A set of thresholds ([$0\!-\!255$]) is applied to {binarize} the prediction. Each threshold produces a pair of precision/recall values to form a PR curve for describing model performance.

\noindent\textbullet~\textbf{F-measure~\cite{achanta2009frequency}} comprehensively considers both precision and recall by computing the weighted harmonic mean: 
\begin{equation}\label{equation:fbeta}\small
  F_{\beta}= \frac{(1+\beta^2)\text{Precision}\times \text{Recall}}{\beta^2 \text{Precision} + \text{Recall}}.
\end{equation}
Empirically, $\beta^2$ is set to $0.3$~\cite{achanta2009frequency} to {put more emphasis} on precision. Instead of plotting the whole F-measure curve, some methods only report \emph{maximal} $F_\beta$, or {binarize} the~predicted saliency map  by an adaptive threshold, \ie, twice the mean value of the saliency prediction, and report \emph{mean F}.

\noindent\textbullet~\textbf{MAE~\cite{perazzi2012saliency}} measures the average pixel-wise absolute error between normalized saliency prediction map $\bm{S}\!\in\![0,1]^{W\!\times\! H}$ and binary ground-truth mask $\bm{G}\!\in\!\{0,1\}^{W\!\times\! H}$:
\begin{equation}\label{equation:mae}\small
  \text{MAE}= \frac{1}{W\!\times\!H} \sum\nolimits_{i=1}^{W} \sum\nolimits_{j=1}^{H} \lvert \bm{G}(i,j)-\bm{S}(i,j) \rvert.
\end{equation}

\begin{table*}
\centering
\caption{Benchmarking results of $44$ state-of-the-art deep SOD models and $3$ top-performing classic SOD methods on $6$ famous datasets (\S\ref{sec:overall_bmk}). Here max~\texttt{F}, \texttt{S}, and \texttt{M} indicate \emph{maximal} $F_\beta$, S-measure, and MAE, respectively.  The three best scores are marked in \textcolor{red}{\textbf{red}}, \textcolor{blue}{\textbf{blue}}, and \textcolor{green}{\textbf{green}}, respectively.}
\begin{threeparttable}
\resizebox{0.99\textwidth}{!}{
\setlength\tabcolsep{2.1pt}
\renewcommand\arraystretch{1.02}
\begin{tabular}{|c|r||c|c|c|c|c|c|c|c|c|c|c|c|c|c|c|c|c|c|}  
\hline\thickhline
\multicolumn{2}{|c||}{Dataset}    &\multicolumn{3}{c|}{ECSSD~\cite{shi2015hierarchical}}
            &\multicolumn{3}{c|}{DUT-OMRON~\cite{yang2013}}
            &\multicolumn{3}{c|}{PASCAL-S~\cite{li2014secrets}}
            &\multicolumn{3}{c|}{HKU-IS~\cite{li2015visual}}
            &\multicolumn{3}{c|}{DUTS\texttt{-test}~\cite{wang2017}}
            &\multicolumn{3}{c|}{SOD~\cite{Movahedi2010}} 
            \\
\hline
\multicolumn{2}{|c||}{Metric}     &max~\texttt{F}$\uparrow$ &~~~\texttt{S}$\uparrow$~~ &~~~\texttt{M}$\downarrow$~~
            &max~\texttt{F}$\uparrow$ &~~~\texttt{S}$\uparrow$~~ &~~~\texttt{M}$\downarrow$~~
            &max~\texttt{F}$\uparrow$ &~~~\texttt{S}$\uparrow$~~ &~~~\texttt{M}$\downarrow$~~
            &max~\texttt{F}$\uparrow$ &~~~\texttt{S}$\uparrow$~~ &~~~\texttt{M}$\downarrow$~~
            &max~\texttt{F}$\uparrow$ &~~~\texttt{S}$\uparrow$~~ &~~~\texttt{M}$\downarrow$~~
            &max~\texttt{F}$\uparrow$ &~~~\texttt{S}$\uparrow$~~ &~~~\texttt{M}$\downarrow$~~
            \\

\hline
\hline
\multirow{3}{*}{\rotatebox{90}{2013-14}}
 &$^*$HS\cite{yan2013hierarchical}&.673 &.685 &.228
                                &.561 &.633 &.227
                                &.569 &.624 &.262
                                &.652 &.674 &.215
                                &.504 &.601 &.243
                                &.756 &.711 &.222
                                \\

 &$^*$DRFI\cite{jiang2013salient} &.751 &.732 &.170
                                &.623 &.696 &.150
                                &.639 &.658 &.207
                                &.745 &.740 &.145
                                &.600 &.676 &.155
                                &.658 &.619 &.228
                                \\
 &$^*$wCtr\cite{zhu2014saliency} &.684 &.714 &.165
                              &.541 &.653 &.171
                              &.599 &.656 &.196
                              &.695 &.729 &.138
                              &.522 &.639 &.176
                              &.615 &.638 &.213
                              \\
\hline
\multirow{3}{*}{\rotatebox{90}{2015}}
 &MCDL\cite{zhao2015saliency}  &.816 &.803 &.101
                                &.670 &.752 &.089
                                &.706 &.721 &.143
                                &.787 &.786 &.092
                                &.634 &.713 &.105
                                &.689 &.651 &.182
                                \\

 &LEGS\cite{wang2015deep} &.805 &.786 &.118
                           &.631 &.714 &.133
                           &$\ddagger$ &$\ddagger$ &$\ddagger$
                           &.736 &.742 &.119
                           &.612 &.696 &.137
                           &.685 &.658 &.197
                           \\

 &MDF\cite{li2015visual}  &.797 &.776 &.105
                           &.643 &.721 &.092
                           &.704 &.696 &.142
                           &.839 &.810 &.129
                           &.657 &.728 &.114
                           &.736 &.674 &.160
                           \\
\hline
\multirow{6}{*}{\rotatebox{90}{2016}}
 &ELD\cite{lee2016deep} &.849 &.841 &.078
                         &.677 &.751 &.091
                         &.782 &.799 &.111
                         &.868 &.868 &.063
                         &.697 &.754 &.092
                         &.717 &.705 &.155
                         \\
 &DHSNet\cite{liu2016dhsnet} &.893 &.884 &.060
                              &$\ddagger$ &$\ddagger$ &$\ddagger$
                              &.799 &.810 &.092
                              &.875 &.870 &.053
                              &.776 &.818 &.067
                              &.790 &.749 &.129
                              \\
 &DCL\cite{li2016deep} &.882 &.868 &.075
                        &.699 &.771 &.086
                        &.787 &.796 &.113
                        &.885 &.877 &.055
                        &.742 &.796 &.149
                        &.786 &.747 &.195
                        \\
 &$^\diamond$MAP\cite{zhang2016unconstrained} &.556 &.611 &.213
                                    &.448 &.598 &.159
                                    &.521 &.593 &.207
                                    &.552 &.624 &.182
                                    &.453 &.583 &.181
                                    &.509 &.557 &.236
                                    \\

 &CRPSD\cite{tang2016saliency} &.915 &.895 &.048
                                &- &- &-
                                &.864 &.852 &.064
                                &.906 &.885 &.043
                                &- &- &-
                                &- &- &-
                                \\
 &RFCN\cite{wang2016saliency}  &.875 &.852 &.107
                                &.707 &.764 &.111
                                &.800 &.798 &.132
                                &.881 &.859 &.089
                                &.755 &.859 &.090
                                &.769 &.794 &.170
                                \\
\hline
\multirow{10}{*}{\rotatebox{90}{2017}}
 &MSRNet\cite{li2017instance}  &.900 &.895 &.054
                                &.746 &.808 &.073
                                &.828 &.838 &.081
                                &$\ddagger$ &$\ddagger$ &$\ddagger$
                                &.804 &.839 &.061
                                &.802 &.779 &.113
                                \\
 &DSS\cite{hou2016deeply}  &.906 &.882 &.052
                              &.737 &.790 &.063
                              &.805 &.798 &.093
                              &$\ddagger$ &$\ddagger$ &$\ddagger$
                              &.796 &.824 &.057
                              &.805 &.751 &.122
                              \\
 &$^\dagger$WSS\cite{wang2017} &.879 &.811 &.104
                                &.725 &.730 &.110
                                &.804 &.744 &.139
                                &.878 &.822 &.079
                                &.878 &.822 &.079
                                &.807 &.675 &.170
                                \\
 &DLS\cite{hu2017deep} &.826 &.806 &.086
                        &.644 &.725 &.090
                        &.712 &.723 &.130
                        &.807 &.799 &.069
                        &- &- &-
                        &- &- &-
                        \\
 &NLDF\cite{luo2017non} &.889 &.875 &.063
                         &.699 &.770 &.080
                         &.795 &.805 &.098
                         &.888 &.879 &.048
                         &.777 &.816 &.065
                         &.808 &.889 &.125 
                          \\
 &Amulet\cite{zhang2017amulet}  &.905 &.894 &.059
                                &.715 &.780 &.098
                                &.805 &.818 &.100
                                &.887 &.886 &.051
                                &.750 &.804 &.085
                                &.773 &.757 &.142
                                \\
 &FSN\cite{chen2017look} &.897 &.884 &.053
                          &.736 &.802 &.066
                          &.800 &.804 &.093
                          &.884 &.877 &.044
                          &.761 &.808 &.066
                          &.781 &.755 &.127
                          \\
 &SBF\cite{zhang2017supervision}&.833 &.832 &.091
                                &.649 &.748 &.110
                                &.726 &.758 &.133
                                &.821 &.829 &.078
                                &.657 &.743 &.109
                                &.740 &.708 &.159
                                 \\
 &SRM\cite{Wang_2017_ICCV} &.905 &.895 &.054
                            &.725 &.798 &.069
                            &.817 &.834 &.084
                            &.893 &.887 &.046
                            &.798 &.836 &.059
                            &.792 &.741 &.128
                            \\
 &UCF\cite{Zhang_2017_ICCV} &.890 &.883 &.069
                             &.698 &.760 &.120
                             &.787 &.805 &.115
                             &.874 &.875 &.062
                             &.742 &.782 &.112
                             &.763 &.753 &.165
                             \\
\hline
\multirow{9}{*}{\rotatebox{90}{2018}}
 &RADF\cite{hu2018recurrently} &.911 &.894 &.049
                                &.761 &.817 &\textcolor{green}{\textbf{.055}}
                                &.800 &.802 &.097
                                &.902 &.888 &.039
                                &.792 &.826 &.061
                                &.804 &.757 &.126
                                \\
 &BDMP\cite{zhang2018bi} &.917 &.911 &.045
                          &.734 &.809 &.064
                          &.830 &.845 &.074
                          &.910 &.907 &.039
                          &.827 &.862 &.049
                          &.806 &.786 &.108
                          \\
 &DGRL\cite{wang2018detect} &.916 &.906 &.043
                             &.741 &.810 &.063
                             &.830 &.839 &.074
                             &.902 &.897 &.037
                             &.805 &.842 &.050 
                             &.802 &.771 &.105
                             \\
 &PAGR\cite{zhang2018progressive} &.904 &.889 &.061
                                    &.707 &.775 &.071
                                    &.814 &.822 &.089
                                    &.897 &.887 &.048
                                    &.817 &.838 &.056
                                    &.761 &.716 &.147
                                    \\
 &RSDNet\cite{Islam_2018_CVPR}      &.880 &.788 &.173
                                    &.715 &.644 &.178
                                    &$\ddagger$ &$\ddagger$ &$\ddagger$
                                    &.871 &.787 &.156
                                    &.798 &.720 &.161
                                    &.790 &.668 &.226
                                 \\
 &ASNet\cite{Wang_2018_CVPR} &.925 &.915 &.047
                             &$\ddagger$ &$\ddagger$ &$\ddagger$
                             &\textcolor{green}{\textbf{.848}} &\textcolor{green}{\textbf{.861}} &.070
                             &.912 &.906 &.041
                             &.806 &.843 &.061
                             &.801 &.762 &.121
                                 \\ 

 &PiCANet\cite{liu2018picanet}  &.929 &.916 &\textcolor{red}{\textbf{.035}}
                                    &.767 &.825 &\textcolor{blue}{\textbf{.054}}
                                    &.838 &.846 &\textcolor{blue}{\textbf{.064}}
                                    &.913 &.905 &\textcolor{blue}{\textbf{.031}}
                                    &.840 &.863 &\textcolor{green}{\textbf{.040}}
                                    &.814 &.776 &\textcolor{red}{\textbf{.096}}
                                    \\ 

 &$^\dagger$C2S-Net\cite{li2018contour} &.902 &.896 &.053
                               &.722 &.799 &.072
                               &.827 &.839 &.081
                               &.887 &.889 &.046
                               &.784 &.831 &.062
                               &.786 &.760 &.124 
                               \\
 &RAS\cite{chen2018reverse}   &.908 &.893 &.056
                               &.753 &.814 &.062
                               &.800 &.799 &.101
                               &.901 &.887 &.045
                               &.807 &.839 &.059
                               &.810 &.764 &.124 
                               \\
\hline

\multirow{16}{*}{\rotatebox{90}{2019}}
%

 &AFNet\cite{feng2019attentive} &.924 &.913 &.042
                                &.759 &.826 &.057
                                &.844 &.849 &.070
                                &.910 &.905 &.036
                                &.838 &.867 &.046
                                &.809 &.774 &.111
                                \\

 &BASNet\cite{qin2019basnet} &.931 &.916 &\textcolor{blue}{\textbf{.037}}
                             &\textcolor{green}{\textbf{.779}} &\textcolor{green}{\textbf{.836}} &.057
                             &.835 &.838 &.076
                             &.919 &.909 &\textcolor{green}{\textbf{.032}}
                             &.838 &.866 &.048 
                             &.805 &.769 &.114
                             \\
 &CapSal\cite{zhang2019capsal} &.813 &.826 &.077
                          &.535 &.674 &.101
                          &.827 &.837 &.073
                          &.842 &.851 &.057
                          &.772 &.818 &.061
                          &.669 &.694 &.148
                          \\

 &CPD\cite{wu2019cascaded} &.926 &.918 &\textcolor{blue}{\textbf{.037}}
                            &.753 &.825 &.056
                            &.833 &.848 &.071
                            &.911 &.905 &.034
                            &.840 &.869 &.043
                            &.814 &.767 &.112
                                    \\
 &MLSLNet\cite{wu2019mutual}      &.917 &.911 &.045
                                    &.734 &.809 &.064
                                    &.835 &.844 &.074
                                    &.910 &.907 &.039
                                    &.828 &.862 &.049
                                    &.806 &.786 &.108
                                 \\
 &$^\dagger$MWS\cite{zeng2019multi} &.859 &.827 &.099
		                            &.676 &.756 &.108
		                            &.753 &.768 &.134
		                            &.835 &.818 &.086
		                            &.720 &.759 &.092
		                            &.772 &.700 &.170
                                 \\ 

 &PAGE-Net\cite{wang2019salient}  &.926 &.910 &\textcolor{blue}{\textbf{.037}}
                                    &.760 &.819 &.059
                                    &.829 &.835 &.073
                                    &.910 &.901 &\textcolor{blue}{\textbf{.031}}
                                    &.816 &.848 &.048
                                    &.795 &.763 &.108
                                    \\ 

 &PS\cite{wang2019iterative} &.930 &.918 &\textcolor{green}{\textbf{.041}}
                               &\textcolor{red}{\textbf{.789}} &\textcolor{blue}{\textbf{.837}} &.061
                               &.837 &.850 &.071
                               &.913 &.907 &.038
                               &.835 &.865 &.048
                               &.824 &.800 &\textcolor{green}{\textbf{.103}} 
                               \\
 &PoolNet\cite{liu2019simple}   &\textcolor{blue}{\textbf{.937}} &\textcolor{blue}{\textbf{.926}} &\textcolor{red}{\textbf{.035}}
                               &.762 &.831 &\textcolor{blue}{\textbf{.054}}
                               &\textcolor{red}{\textbf{.858}} &\textcolor{blue}{\textbf{.865}} &\textcolor{green}{\textbf{.065}}
                               &\textcolor{blue}{\textbf{.923}} &\textcolor{red}{\textbf{.919}} &\textcolor{red}{\textbf{.030}}
                               &\textcolor{blue}{\textbf{.865}} &\textcolor{blue}{\textbf{.886}} &\textcolor{red}{\textbf{.037}}
                               &\textcolor{green}{\textbf{.831}} &\textcolor{green}{\textbf{.788}} &.106 
                               \\

 &BANet-R\cite{su2019selectivity}   &\textcolor{red}{\textbf{.939}} &.924 &\textcolor{red}{\textbf{.035}}
                               &\textcolor{blue}{\textbf{.782}} &.832 &.059
                               &.847 &.852 &.070
                               &\textcolor{blue}{\textbf{.923}} &.913 &\textcolor{green}{\textbf{.032}}
                               &.858 &.879 &.040
                               &\textcolor{blue}{\textbf{.842}} &\textcolor{blue}{\textbf{.791}} &.106 
                               \\

 &EGNet-R\cite{zhao2019egnet}  &\textcolor{green}{\textbf{.936}} &\textcolor{green}{\textbf{.925}} &\textcolor{blue}{\textbf{.037}}
                               &.777 &\textcolor{red}{\textbf{.841}} &\textcolor{red}{\textbf{.053}}
                               &.841 &.852 &.074
                               &\textcolor{red}{\textbf{.924}} &\textcolor{blue}{\textbf{.918}} &\textcolor{blue}{\textbf{.031}}
                               &\textcolor{red}{\textbf{.866}} &\textcolor{red}{\textbf{.887}} &\textcolor{blue}{\textbf{.039}}
                               &\textcolor{red}{\textbf{.854}} &\textcolor{red}{\textbf{.802}} &\textcolor{blue}{\textbf{.099}} 
                               \\

 &HRSOD-DH\cite{zeng2019towards}   &.911 &.888 &.052
                               &.692 &.762 &.065
                               &.810 &.817 &.079
                               &.890 &.877 &.042
                               &.800 &.824 &.050
                               &.735 &.705 &.139
                               \\

 &JDFPR\cite{xu2019structured}  &.915 &.907 &.049
                               &.755 &.821 &.057
                               &.827 &.841 &.082
                               &.905 &.903 &.039
                               &.792 &.836 &.059
                               &.792 &.763 &.123 
                               \\

 &SCRN\cite{wu2019stacked}   &\textcolor{blue}{\textbf{.937}} &\textcolor{red}{\textbf{.927}} &\textcolor{blue}{\textbf{.037}}
                               &.772 &\textcolor{green}{\textbf{.836}} &.056
                               &\textcolor{blue}{\textbf{.856}} &\textcolor{red}{\textbf{.869}} &\textcolor{red}{\textbf{.063}}
                               &\textcolor{green}{\textbf{.921}} &\textcolor{green}{\textbf{.916}} &.034
                               &\textcolor{green}{\textbf{.864}} &\textcolor{green}{\textbf{.885}} &\textcolor{green}{\textbf{.040}}
                               &.826 &.787 &.107 
                               \\

&SSNet\cite{zeng2019joint}  &.889 &.867 &.046
                               &.708 &.773 &.056
                               &.793 &.807 &.072
                               &.876 &.854 &.041
                               &.769 &.784 &.049
                               &.713 &.700 &.118 
                               \\
&TSPOANet\cite{liu2019employing}   &.919 &.907 &.047
                               &.749 &.818 &.061
                               &.830 &.842 &.078
                               &.909 &.902 &.039
                               &.828 &.860 &.049
                               &.810 &.772 &.118 
                               \\
\hline
\end{tabular}
}
\begin{tablenotes}
\item[]$^*$ {\scriptsize Non-deep learning model.} $^\dagger$ {\scriptsize Weakly-supervised model.} $^\diamond$ {\scriptsize Bounding-box output.} $\ddagger$ {\scriptsize Training on subset.} - {\scriptsize Results not available.}
\end{tablenotes}
\end{threeparttable}
\label{table:benchmark_all}
\end{table*}
\noindent\textbullet~\textbf{Fbw~\cite{margolin2014evaluate}} intuitively generalizes F-measure by alternating the way {of calculating} precision and recall. It extends the four basic quantities TP, TN, FP and FN to real values, and assigns different weights ($\omega$) to different errors at different locations, considering the neighborhood information:
\begin{equation}\label{equation:fbeta_omg}\small
  F_\beta^\omega= \frac{(1+\beta^2)\text{Precision}^\omega\times \text{Recall}^\omega}{\beta^2 \text{Precision}^\omega + \text{Recall}^\omega}.
\end{equation}

\noindent\textbullet~\textbf{S-measure~\cite{FanStructMeasureICCV17}} evaluates the structural similarity between the real-valued saliency map and the binary ground-truth. It considers object-aware ($S_o$) and region-aware ($S_r$) structure similarities:
\begin{equation}\label{equation:s_measure}\small
  S= \alpha \times S_o + (1-\alpha)\times S_r,
\end{equation}
where $\alpha$ is empirically set to $0.5$.

\noindent\textbullet~\textbf{E-measure~\cite{Fan2018Enhanced}} considers global means of the image and local pixel matching simultaneously:
\begin{equation}\label{equation:e_measure}\small
  Q_{\bm{S}}= \frac{1}{W\!\times\!H} \sum\nolimits_{i=1}^{W} \sum\nolimits_{j=1}^{H} \phi_{\bm{S}}(i,j),
\end{equation}
where $\phi_{\bm{S}}$ is the enhanced alignment matrix, reflecting the correlation between $\bm{S}$ and $\bm{G}$ after subtracting their global means, respectively.

\subsection{{Discussion}}\label{sec:medis}
{These measures are typically based
on \textit{pixel-wise} errors while ignoring$_{\!}$ \textit{structural}$_{\!}$ similarities, with S-measure$_{\!}$ and E-measure {being the only exceptions}. F-measure and E-measure are designed for assessing \textit{binarized} saliency prediction maps, while PR, MAE, Fbw, and S-measure are for \textit{non-binary} map evaluation.}

{Among pixel-level metrics, the PR curve is classic. {However,} precision and recall cannot fully assess the quality of saliency predictions, since high-precision predictions may only highlight a part of salient objects, {while} high-recall {predictions are typically} meaningless if all the pixels are predicted as {being} salient. In general, a high-recall response may come at the expense of reduced precision, and vice versa. F-measure and Fbw are thus used to consider precision and recall simultaneously. However, overlap-based metrics (\ie, PR, F-measure, and Fbw) do not consider the true negative saliency assignments, \ie, the pixels correctly
marked as non-salient. Thus, these metrics favor methods that successfully assign
high saliency to salient pixels but fail to detect non-salient regions~\cite{borji2014salient}.  MAE can remedy~this, but it performs poorly when salient objects are small.  For the structure-/image-level metrics, S-measure is more popular than E-measure, as SOD focuses on continuous predictions.}

{Considering$_{\!}$ the$_{\!}$ popularity$_{\!}$ and$_{\!}$ characteristics$_{\!}$ of$_{\!}$ existing$_{\!}$ metrics$_{\!}$ and$_{\!}$ completeness$_{\!}$ of$_{\!}$ evaluation, F-measure (\emph{maximal}$_{\!}$ $F_\beta$), S-measure$_{\!}$ and$_{\!}$ MAE$_{\!}$ are$_{\!}$ our$_{\!}$ top$_{\!}$ recommendations.}

\begin{figure*}[t]
  \centering
      \includegraphics[width=1\linewidth]{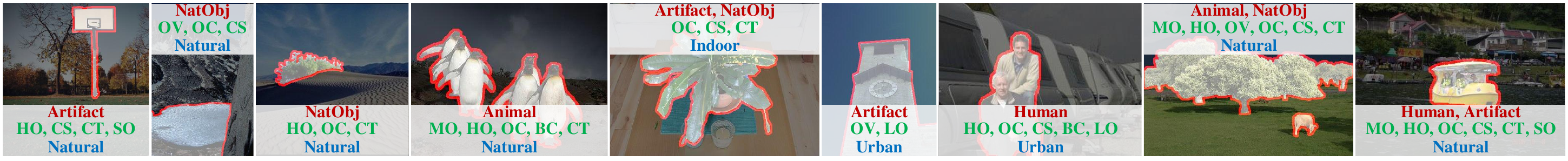}
\caption{Sample images from the hybrid benchmark consisting of images randomly selected from $6$ SOD datasets. Salient regions are uniformly highlighted. Corresponding attributes are listed. See \S\ref{sec:attribute_eval} for more detailed descriptions.}
\label{fig:Hybrid_bmk}
\end{figure*}

\section{Benchmarking and Empirical Analysis}
\label{sec:benchmarking}
{This section provides empirical analyses to shed light on some key challenges in the field. Specifically, with our large-scale benchmarking (\S\ref{sec:overall_bmk}), we first conduct an attribute-based study to better understand the benefits and limitations of$_{\!}$
current$_{\!}$ {arts}$_{\!}$ (\S\ref{sec:attribute_eval}). Then, we$_{\!}$ study$_{\!}$ the$_{\!}$ robustness$_{\!}$ of SOD$_{\!}$ models$_{\!}$ against$_{\!}$ input$_{\!}$ perturbations, \ie, random$_{\!}$ exerted\!\! noises$_{\!}$ (\S\ref{sec:input_ptb}) and$_{\!}$ manually designed adversarial samples$_{\!}$ (\S\ref{sec:adversarial_atk}). Finally, we$_{\!}$  quantitatively$_{\!}$  assess$_{\!}$  the$_{\!}$  generalizability$_{\!}$  and$_{\!}$  {difficulty}$_{\!}$  of$_{\!}$  current$_{\!}$  mainstream$_{\!}$  SOD datasets$_{\!}$ (\S\ref{sec:cd_generalization}).}

\begin{table*}[t]
\renewcommand\thetable{7}
\centering
\caption{Attribute-based study \wrt salient object categories, challenges and scene categories. $(\cdot)$ indicates the percentage of images with a specific attribute.
\textit{ND-avg} indicates the average score of three heuristic models: HS~\cite{yan2013hierarchical}, DRFI~\cite{jiang2013salient} and wCtr~\cite{zhu2014saliency}.
\textit{D-avg} indicates the average score of three deep learning models: DGRL~\cite{wang2018detect}, PAGR~\cite{zhang2018progressive} and PiCANet~\cite{liu2018picanet}. Best in \textcolor{red}{\textbf{red}}, and worst with \underline{\textbf{underline}}. See \S\ref{sec:attribute_eval} for more details.}
\renewcommand\arraystretch{1.05}
\begin{threeparttable}
\resizebox{0.99\textwidth}{!}{
\setlength\tabcolsep{4pt}
\begin{tabular}{|c||r||c|c|c|c||c|c|c|c|c|c|c|c|c||c|c|c|}
\hline\thickhline
\multirow{3}*{Metric} &\multirow{3}*{Method~~~~}
&\multicolumn{4}{c||}{Salient object categories}
&\multicolumn{9}{c||}{Challenges}
&\multicolumn{3}{c|}{Scene categories}
\\
\cline{3-18}

&
&\textit{Human} &\textit{Animal} &\textit{Artifact} &\textit{NatObj}
&$\mathcal{MO}$ &$\mathcal{HO}$ &$\mathcal{OV}$ &$\mathcal{OC}$ &$\mathcal{CS}$ &$\mathcal{BC}$ &$\mathcal{CT}$ &$\mathcal{SO}$ &$\mathcal{LO}$ 
&\textit{Indoor} &\textit{Urban} &\textit{Natural}\\

&
&(26.61) &(38.44) &(45.67) &(10.56)
&(11.39) &(66.39) &(28.72) &(46.50) &(40.44) &(47.22) &(74.11) &(21.61) &(12.61) 
&(20.28) &(22.22) &(57.50) \\
\hline
\hline
\multirow{8}*{max~\texttt{F}$\uparrow$}
&$^*$HS\cite{yan2013hierarchical}
    &.587 &.650 &.636 &.704
    &.663 &.637 &.631 &.645 &.558 &.647 &.629 &.493 &.737 
    &.594 &.627 &.650 \\
&$^*$DRFI\cite{jiang2013salient}
    &.635 &.692 &.673 &.713
    &.674 &.688 &.658 &.675 &.599 &.662 &.677 &.566 &.747 
    &.609 &.661 &.697 \\
&$^*$wCtr\cite{zhu2014saliency}
    &.557 &.621 &.624 &.682
    &.639 &.625 &.605 &.620 &.522 &.612 &.606 &.469 &.689 
    &.578 &.613 &.618 \\
&DGRL\cite{wang2018detect}
    &.820 &.881 &.830 &.728
    &.783 &.846 &.829 &.830 &.781 &.842 &.834 &.724 &.873 
    &.800 &.848 &.840 \\
&PAGR\cite{zhang2018progressive}
    &.834 &.890 &.787 &.725
    &.743 &.819 &.778 &.809 &.770 &.797 &.822 &.760 &.802 
    &.788 &.796 &.828 \\
&PiCANet\cite{liu2018picanet}
    &.840 &.897 &.846 &.669
    &.791 &.861 &.843 &.845 &.797 &.848 &.850 &.763 &.889 
    &.806 &.862 &.859 \\
\cline{2-18}
&\textit{$^*$ND-avg}
    &\underline{\textbf{.593}} &.654 &.644 &\cred \textbf{.700}
    &.659 &.650 &.631 &.647 &.560 &.640 &.637 &\underline{\textbf{.509}} &\cred \textbf{.724} 
    &\underline{\textbf{.594}} &.634 &\cred \textbf{.655} \\
&\textit{D-avg}
    &.831 &\cred \textbf{.889} &.821 &\underline{\textbf{.708}}
    &.772 &.842 &.817 &.828 &.783 &.829 &.836 &\underline{\textbf{.749}} &\cred \textbf{.855} 
    &\underline{\textbf{.798}} &.836 &\cred \textbf{.842} \\
\hline

\end{tabular}
}
\begin{tablenotes}
\footnotesize
\item[]$^*$ {\scriptsize Non-deep learning model.}
\end{tablenotes}
\end{threeparttable}
\label{table:att_stat}
\end{table*}

\subsection{{Quick Overview}}
\label{sec:benchmark_quick}

{For ease of understanding, we {compile important} observations and conclusions from subsequent experiments below.}

\noindent{\textbullet~\textit{Overall benchmarks}~(\S\ref{sec:overall_bmk}). As shown in Table~\ref{table:benchmark_all}, deep SOD models significantly outperform
heuristic ones, and the performance on some datasets~\cite{shi2015hierarchical,li2015visual} {has become} saturated. \cite{liu2019simple,su2019selectivity,zhao2019egnet,wu2019stacked} are current state-of-the-arts.}

\noindent{\textbullet~\textit{Attribute-based analysis}~(\S\ref{sec:attribute_eval}).
Results in Table~\ref{table:att_stat} reveal that deep methods show significant advantages in {detecting} semantic-rich objects, such as animal. Both deep and non-deep methods face difficulties with small salient objects. For application scenarios, indoor scenes pose great challenges, {highlighting} potential directions for future efforts. }

\noindent{\textbullet~\textit{Robustness against random perturbations}~(\S\ref{sec:input_ptb}). As shown in Table~\ref{table:input_p},  surprisingly,  deep methods are more sensitive than heuristic ones to random input perturbations. Both types of methods {demonstrate more robustness} against \textit{Rotation}, while being fragile towards \textit{Gaussian blur} and \textit{Gaussian noise}.}

\noindent{\textbullet~\textit{Adversarial attack}~(\S\ref{sec:adversarial_atk}). Table~\ref{table:adv_atk} suggests that adversarial attacks cause drastic {degradation in} performance for deep SOD models, {and are} even worse than that of random perturbations. {However, attacks} rarely transfers {between} different SOD networks.}

\noindent{\textbullet~\textit{Generalizability and {difficulty} of datasets}~(\S\ref{sec:cd_generalization}). Table~\ref{table:cdg} {shows} that
DUTS-\textit{train}~\cite{wang2017} is a good choice for training deep SOD models as it has the best generalizability, while
SOC~\cite{Fan_2018_ECCV}, DUT-OMRON~\cite{yang2013}, and DUTS-\textit{test}~\cite{wang2017} are more suitable for evaluation due to their {difficulty}.
}

\begin{table}[t]
\centering
\renewcommand\thetable{6}
\caption{
         Descriptions of attributes {that often bring difficulties to SOD (see \S\ref{sec:attribute_eval})}.}
\renewcommand\arraystretch{1.03}
\begin{threeparttable}
\resizebox{0.49\textwidth}{!}{
\setlength\tabcolsep{2pt}
\begin{tabular}{l l}
\hline\thickhline
Attr &~~~~~~~~~~~~~~~~~~~~~~~~~~~~~~~~~~~~~~~~~Description \\
\hline
$\mathcal{MO}$ &\textbf{Multiple Objects.} There exist more than two salient objects.\\
$\mathcal{HO}$ &\textbf{Heterogeneous Object.} Salient object regions have distinct \\&colors or illuminations.\\
$\mathcal{OV}$ &\textbf{Out-of-View.} Salient objects are partially clipped by image \\&boundaries.\\
$\mathcal{OC}$ &\textbf{Occlusion.} Salient objects are occluded by other objects.\\
$\mathcal{CS}$ &\textbf{Complex Scene.} Background regions contain confusing \\&objects or rich details.\\
$\mathcal{BC}$ &\textbf{Background Clutter.} Foreground and background regions \\&around the salient object  boundaries have similar colors ($\chi^2$ \\&between RGB histograms less than $0.9$).\\
$\mathcal{CT}$ &\textbf{Complex Topology.} Salient objects have complex shapes, \eg, \\&thin parts or holes.\\
$\mathcal{SO}$ &\textbf{Small Object.} Ratio between salient object area and image is \\& less than $0.1$.\\
$\mathcal{LO}$ &\textbf{Large Object.} Ratio between salient object area and image is \\& larger than $0.5$.\\
\hline
\end{tabular}
}
\end{threeparttable}
\label{table:att_desp}
\end{table}

\subsection{Performance Benchmarking}
\label{sec:overall_bmk}
Table~\ref{table:benchmark_all} shows the performances of $44$ state-of-the-art deep SOD models and three top-performing classic methods (suggested by~\cite{borji2015salient}) on six most popular modern datasets. The performance is measured by three metrics, \ie, \emph{maximal} $F_\beta$, S-measure and MAE, as recommended in \S\ref{sec:medis}. All the benchmarked models are representative, and have publicly available implementations or saliency prediction results. {For performance benchmarking, we either use saliency maps provided by the authors or run their official codes. It is worth mentioning that, for some methods, our benchmarking results are inconsistent with their reported scores. There are several reasons. First, our community long lacked an open, universally-adopted evaluation tool, while there are many implementation factors would influence the evaluation scores, such as input image resolution, threshold step, \etc. Second, some methods~\cite{Zhang_2017_ICCV,wang2018detect,zhang2017amulet,zhang2019capsal,li2019supervae,zhuge2019deep} use \emph{mean} F-measure instead of \emph{maximal} F-measure for performance evaluation. Third, for some methods~\cite{zhang2017amulet,hou2016deeply}, the evaluation scores of finally released saliency maps are inconsistent with the ones reported in papers. We hope that our performance benchmarking, publicly released evaluation tools and SOD maps could help our community build an open and standardized evaluation system and ensure consistency and procedural correctness for results and conclusions produced by different parties.}

Not surprisingly, data-driven models greatly
outperform conventional heuristic ones, due to their strong learning ability for visually salient pattern modeling.  In addition, the performance has gradually increased since 2015, demonstrating well the advancement of deep learning techniques. However, after 2018, {the rate of improvement began decrasing}, calling for more effective model designs and new machine learning technologies. We also find that the performances tend to be saturated on older SOD datasets such as ECSSD~\cite{shi2015hierarchical} and HKU-IS~\cite{li2015visual}.
Hence, among the $44$ famous deep SOD models, we would like to nominate PoolNet~\cite{liu2019simple}, BANet~\cite{su2019selectivity}, EGNet~\cite{zhao2019egnet}, and SCRN~\cite{wu2019stacked} as the four state-of-the-art methods, which consistently show promising performance over diverse datasets.

\subsection{Attribute-Based Study}
\label{sec:attribute_eval}
Although the community has witnessed the great advances made by deep SOD models, it is still unclear {under which} specific aspects these models perform well. As there are numerous factors affecting the performance of a SOD algorithm, such as object/scene category, occlusion, \etc, it is crucial to evaluate the performance under {different} scenarios. This can help {reveal the} strengths and weaknesses of deep SOD models, identify pending challenges, and {highlight} future research directions towards more robust algorithms.

\subsubsection{Hybrid Benchmark Dataset with Attribute Annotations}
\begin{table*}[t]
\centering
\caption{Attribute statistics of top and bottom 100 images based on F-measure.
$(\cdot)$ indicates the percentage of the images with a specific attribute.
\textit{ND-avg} indicates the average results of three heuristic models: HS~\cite{yan2013hierarchical}, DRFI~\cite{jiang2013salient} and wCtr~\cite{zhu2014saliency}.
\textit{D-avg} indicates the average results of three deep models: DGRL~\cite{wang2018detect}, PAGR~\cite{zhang2018progressive} and PiCANet~\cite{liu2018picanet}.
Two largest changes in \textcolor{red}{\textbf{red}} if positive, and \textcolor{blue}{\textbf{blue}} if negative. See \S\ref{sec:attribute_eval} for more details.}
\vspace{-5pt}
\renewcommand\arraystretch{1.05}
\begin{threeparttable}
\resizebox{0.99\textwidth}{!}{
\setlength\tabcolsep{3.8pt}
\begin{tabular}{|c||r||c|c|c|c||c|c|c|c|c|c|c|c|c||c|c|c|}  
\hline\thickhline
\multirow{3}*{Method} &\multirow{3}*{Cases~~~~}
&\multicolumn{4}{c||}{Salient object categories}
&\multicolumn{9}{c||}{Challenges}
&\multicolumn{3}{c|}{Scene categories}
\\
\cline{3-18}

&
&\textit{Human} &\textit{Animal} &\textit{Artifact} &\textit{NatObj} 
&$\mathcal{MO}$ &$\mathcal{HO}$ &$\mathcal{OV}$ &$\mathcal{OC}$ &$\mathcal{CS}$ &$\mathcal{BC}$ &$\mathcal{CT}$ &$\mathcal{SO}$ &$\mathcal{LO}$ 
&\textit{Indoor} &\textit{Urban} &\textit{Natural}\\

&
&(26.61) &(38.44) &(45.67) &(10.56)
&(11.39) &(66.39) &(28.72) &(46.50) &(40.44) &(47.22) &(74.11) &(21.61) &(12.61) 
&(20.28) &(22.22) &(57.50) \\

\hline
\hline
\multirow{4}*{\textit{ND-avg}}
&Best ($\%$) &13.00 &25.00 &46.00 &27.00 
             &5.00 &61.00 &12.00 &26.00 &10.00 &20.00 &63.00 &5.00 &18.00 
             &17.00 &6.00 &12.00 \\
&\textit{change} &\cblue \textbf{-13.61} &-13.44 &+0.33 &\cred \textbf{+14.44}
               &-6.39 &-5.39 &-16.72 &-20.50 &\cblue \textbf{-30.44} &{-27.22} &-11.11 &-16.61 &\cred \textbf{+5.39} 
               &\cred \textbf{-3.28} & {-16.22} &\cblue \textbf{-45.50} \\
\cline{2-18}
&Worst ($\%$)  &36.00 &30.00 &41.00 &5.00 
              &6.00 &54.00 &15.00 &34.00 &70.00 &31.00 &71.00 &76.00 &0.00 
              &22.00 &37.00 &37.00 \\
&\textit{change}  &\cred \textbf{+9.39} &\cblue \textbf{-8.44} &-4.67 &-5.56
               &-5.39 &-12.39 &-13.72 &-12.50 &{+29.56} &\cblue \textbf{-16.22} &-3.11 &\cred \textbf{+54.39} &-12.61 
               &+1.72 &\cred \textbf{+14.78} &\cblue \textbf{-20.50} \\

\hline
\multirow{4}*{\textit{D-avg}}
&Best ($\%$)  &24.00 &30.00 &49.00 &17.00 
             &3.00 &69.00 &33.00 &28.00 &26.00 &35.00 &49.00 &2.00 &18.00 
             &24.00 &23.00 &53.00 \\
&\textit{change}  &-2.61 &\cblue \textbf{-8.44} &+3.33 &\cred \textbf{+6.44}
               &-8.39 &+2.61 &+4.28 &-18.50 &-14.44 &-12.22 &\cblue \textbf{-25.11} &{-19.61} &\cred \textbf{+5.39} 
               &\cred \textbf{+3.72} &+0.78 &\cblue \textbf{-4.50} \\
\cline{2-18}
&Worst ($\%$)  &30.00 &10.00 &49.00 &33.00 
              &20.00 &52.00 &28.00 &46.00 &70.00 &42.00 &59.00 &50.00 &3.00 
              &32.00 &23.00 &45.00 \\
&\textit{change}  &+3.39 &\cblue \textbf{-28.44} &+3.33 &\cred \textbf{+22.44}
               &+8.61 &-14.39 &-0.72 &-0.50 &\cred \textbf{+29.56} &-5.22 &\cblue \textbf{-15.11} &{+28.39} &-9.61 
               &\cred \textbf{+11.72} &+0.78 &\cblue \textbf{-12.50} \\

\hline
\end{tabular}
}
\end{threeparttable}
\label{table:att_best_worst}
\end{table*}

To enable a deeper
analysis and understanding of the performance of an algorithm, it is essential to identify the key
factors and circumstances {influencing} it~\cite{perazzi2016benchmark}. To this end, we construct a \textit{hybrid benchmark} with rich attribute annotations. It consists of 1,800 images randomly selected from six SOD datasets (300 for each), namely SOD~\cite{Movahedi2010}, ECSSD~\cite{shi2015hierarchical}, DUT-OMRON~\cite{yang2013}, PASCAL-S~\cite{li2014secrets}, HKU-IS~\cite{li2015visual} and DUTS \textit{test} set~\cite{wang2017}. Inspired by~\cite{li2014secrets,perazzi2016benchmark}, we annotate each image with an extensive set of attributes covering  typical object types, challenging factors and diverse scene categories. Specifically, the annotated \textbf{salient objects} are categorized into \textit{Human}, \textit{Animal}, \textit{Artifact} and \textit{NatObj} (Natural Objects), where \textit{NatObj} includes natural objects such as fruit, plant, mountains, icebergs, lakes, \etc. The \textbf{challenging factors} describe specific situations that often bring difficulties to SOD, such as occlusions, background clutter, and complex shapes (see Table~\ref{table:att_desp}).
The {image \textbf{scene}s include} \textit{Indoor}, \textit{Urban} and \textit{Natural}, where the last two indicate different outdoor environments.
It is worth mentioning that the attributes are not mutually exclusive. Some sample images with attribute annotations are shown in Fig.~\ref{fig:Hybrid_bmk}. Please note that this benchmark will also be used in \S\ref{sec:input_ptb} and \S\ref{sec:adversarial_atk}.

For the baselines {in our attribute-based analysis}, we choose the three top-performing heuristic models again, \ie, HS~\cite{yan2013hierarchical}, DRFI~\cite{jiang2013salient} and wCtr~\cite{zhu2014saliency}, and three recent famous deep methods, \ie, DGRL~\cite{wang2018detect}, PAGR~\cite{zhang2018progressive} and PiCANet~\cite{liu2018picanet}. All three deep models are trained on DUTS-\textit{train}~\cite{wang2017} and have publicly released implementations.

\subsubsection{Analysis}
In Table~\ref{table:att_stat}, we report the performance on subsets
of our hybrid dataset characterized by a particular attribute. To {provide better} insight, in Table~\ref{table:att_best_worst},
we select images with the best-100 and worst-100 model predictions, and compare the portion distributions of attributes \wrt the ones over the whole dataset.
Below are some {important} observations drawn from these experiments.

\noindent\textbullet~\textbf{`Easy' and `hard' object categories.}
Deep and non-deep SOD  models view object categories differently (Table~\ref{table:att_stat}).
For the deep methods (\textit{D-avg}), \textit{NatObj} is clearly the most challenging one which is probably due to its small number of training samples and complex topologies. \textit{Animal} appears to be the easiest, which can be attributed to its significant semantics.
By contrast, traditional methods (\textit{ND-avg}) {struggle with} \textit{Human}, revealing their limitations in capturing high-level semantics. We are surprised to find that the deep models significantly outperform the non-deep ones over almost all the object categories, except \textit{NatObj}. This demonstrates the value of heuristic assumptions in certain scenes and the potential of embedding human prior knowledge into current deep learning schemes.

\noindent\textbullet~\textbf{Most and least challenging factors.}
Table~\ref{table:att_stat} shows that, interestingly, both deep and non-deep methods handle $\mathcal{LO}$ well. In addition, both types of methods face difficulties with $\mathcal{SO}$, {highlighting} a promising direction for future efforts. Besides, we find that $\mathcal{CS}$ and $\mathcal{MO}$ are challenging for deep models, showing that current solutions still fall short at {determining} the relative importance {of different} objects.

\noindent\textbullet~\textbf{Most and least difficult scenes.}
Deep and heuristic methods perform similarly when {faced with} different scenes (Table~\ref{table:att_stat}). For both types of methods, \textit{Natural} is the easiest, which is reasonable as the scenes are typically simple. {Further, though both contain numerous} objects, \textit{Indoor} is {more challenging} than \textit{Urban} as it often suffers from highly unevenly distributed illumination and more complex scenes. Our experiments also show that the utility of SOD models in real, and especially complex, environments is still limited.

\noindent\textbullet~\textbf{Additional advantages of deep models.}
As shown~in Table~\ref{table:att_stat}, deep models achieve great improvements on semantically rich objects (\textit{Human}, \textit{Animal} and \textit{Artifact}), {demonstrating} advantages in semantic modeling. This is verified again by their good performance on complex object shapes ($\mathcal{HO}$, $\mathcal{OV}$, $\mathcal{OC}$, $\mathcal{CT}$). Deep models also narrow the gap between different scene categories (\textit{Indoor} \textit{v.s.} \textit{Natural}), {indicating an} improved robustness against various backgrounds.

\noindent\textbullet~\textbf{Best and worst predictions.}
From Table~\ref{table:att_best_worst}, in addition to similar conclusions drawn from Table~\ref{table:att_stat}, some unique {and interesting observations can be made}. First, for deep methods, \textit{NatObj} spans a large range {of challenge, containing} both the simplest and hardest samples. Thus, future efforts should pay more attention to the hard samples in \textit{NatObj}. In addition, after considering data distribution bias, $\mathcal{CS}$ is the most challenging factor for deep models.

\begin{figure*}[t]
  \centering
      \includegraphics[width=1 \linewidth]{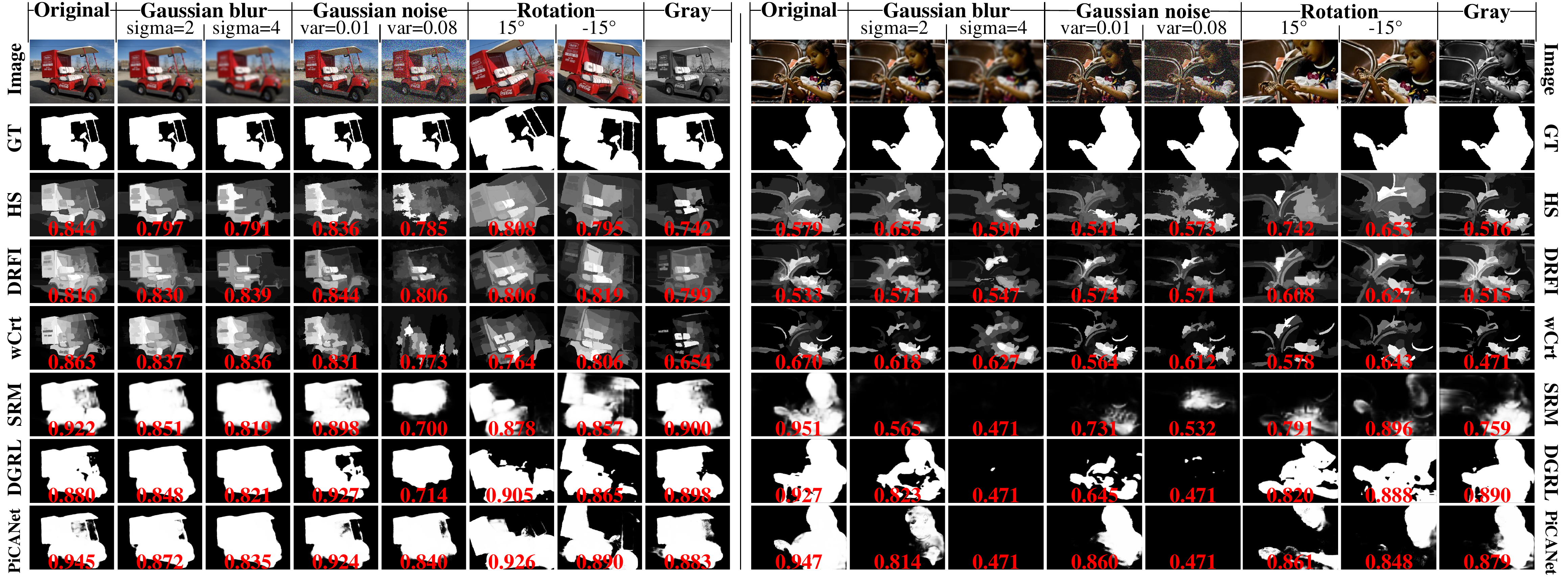}
\caption{Examples of saliency prediction under various input perturbations. The max~\texttt{F} values are denoted in \textcolor{red}{\textbf{red}}. See \S\ref{sec:input_ptb} for more details.}
\label{fig:Input_p}
\end{figure*}

\begin{table}[t]
\centering
\caption{Input perturbation study on the \textbf{hybrid benchmark}.
\textit{ND-avg} indicates the average score of three heuristic models: HS~\cite{yan2013hierarchical}, DRFI~\cite{jiang2013salient} and wCtr~\cite{zhu2014saliency}.
\textit{D-avg} indicates the average score of three deep learning models: SRM~\cite{Wang_2017_ICCV}, DGRL~\cite{wang2018detect} and PiCANet~\cite{liu2018picanet}. Best in \textcolor{red}{\textbf{red}} and worst with \underline{\textbf{underline}}. See \S\ref{sec:input_ptb} for more details. }
\renewcommand\arraystretch{1.05}
\begin{threeparttable}
\resizebox{0.49\textwidth}{!}{
\setlength\tabcolsep{2.5pt}
\begin{tabular}{|c||r||c|c|c|c|c|c|c|c|}  
\hline\thickhline
\multirow{2}*{Metric} &\multirow{2}*{Method~~~~}
&\multirow{2}*{\textit{Original}}&\multicolumn{2}{c|}{\tabincell{c}{\textit{Gaus. blur}\\ ( $\sigma$= )}}
&\multicolumn{2}{c|}{\tabincell{c}{\textit{Gaus. noise}\\ ( var= )}} &\multicolumn{2}{c|}{\textit{Rotation}} &\multirow{2}*{~\textit{Gray}~}
\\
\cline{4-9}
&&&~~~$2$~~~ &$4$ &~$0.01$~ &~$0.08$~ &~~$15^{\circ}$~ &$-15^{\circ}$ &
\\
\hline
\hline
\multirow{8}*{\text{max}~\texttt{F}$\uparrow$}
&$^*$HS\cite{yan2013hierarchical}
    &.600 &-.012 &-.096 &-.022 &-.057 &+.015 &+.009 &-.104 \\
&$^*$DRFI\cite{jiang2013salient}
    &.670 &-.040 &-.103 &-.035 &-.120 &-.009 &-.009 &-.086 \\
&$^*$wCtr\cite{zhu2014saliency}
    &.611 &+.006 &-.000 &-.024 &-.136 &-.004 &-.003 &-.070 \\
&SRM~\cite{Wang_2017_ICCV}
     &.817 &-.090 &-.229 &-.025 &-.297 &-.028 &-.029 &-.042 \\
&DGRL~\cite{wang2018detect}
     &.831 &-.088 &-.365 &-.050 &-.402 &-.031 &-.022 &-.026 \\
&PiCANet~\cite{liu2018picanet}
     &.848 &-.048 &-.175 &-.014 &-.148 &-.005 &-.008 &-.039 \\
\cline{2-10}
&\textit{$^*$ND-avg}
    &.627 &-.015 &-.066 &-.027 &\underline{\textbf{-.104}} &\cred \textbf{-.000} &-.001 &-.087 \\
&\textit{D-avg}
    &.832 &-.075 &-.256 &-.041 &\underline{\textbf{-.282}} &-.021 &\cred \textbf{-.020} &-.037 \\
\hline
\end{tabular}
}
\begin{tablenotes}
\item[]$^*$ {\footnotesize Non-deep learning model.}
\end{tablenotes}
\end{threeparttable}
\label{table:input_p}
\end{table}

\subsection{Robustness {Against} General Input Perturbations}
\label{sec:input_ptb}
The robustness of a model lies in its stability against corrupt inputs.  Intuitively, the outputs of a robust SOD model should be repeatable on slightly different images with the same content. {However}, the recently {introduced} adversarial examples, \ie maliciously constructed inputs that fool machine learning models, can degrade the performance of deep image classifiers significantly. Current deep SOD models {likely} face a similar challenge. Therefore, in this section, we examine the robustness of SOD models by comparing their outputs {for} randomly perturbed inputs, such as noisy or blurred images. Then, in \S\ref{sec:adversarial_atk}, we will study the robustness to manually designed adversarial examples.

The input perturbations {investigated} include \textit{Gaussian blur}, \textit{Gaussian noise}, \textit{Rotation}, and \textit{Gray}.
For blurring, we employ Gaussian blur kernels with a sigma of $2$ or $4$. For noise, we select two variance values, \ie, $0.01$ and $0.08$, to cover both tiny and medium magnitudes. For rotation, we rotate the images by $+15^{\circ}$ and $-15^{\circ}$, respectively, and cut out the largest box with the original aspect ratio. The gray images are generated using the Matlab \texttt{rgb2gray} function.

As in \S\ref{sec:attribute_eval}, we include three popular heuristic models~\cite{yan2013hierarchical,jiang2013salient,zhu2014saliency} and three deep methods~\cite{Wang_2017_ICCV,wang2018detect,liu2018picanet} in our experiments. Table~\ref{table:input_p} shows the results.
Overall, compared with deep models, heuristic methods are less sensitive towards input perturbations. The compactness and abstractness of superpixels likely
explains much of this. Specifically, heuristic methods are rarely affected by \textit{Rotation}, but perform worse under strong \textit{Gaussian blur}, strong \textit{Gaussian noise} and \textit{Gray}. Deep methods suffer the most under \textit{Gaussian blur} and strong \textit{Gaussian noise}, which may be caused by the damage to shallow-layer features. Deep methods are relatively robust against \textit{Rotation}, revealing the rotation invariance of DNNs brought by the pooling operation.  Interestingly, we further find that, among the three deep models, PiCANet~\cite{liu2018picanet} {demonstrates excellent} robustness against a wide range of input perturbations, including \textit{Gaussian blur}, \textit{Gaussian noise}, and \textit{Rotation}. We attribute this to its effective non-local operation. This reveals that effective network designs can improve the robustness to random perturbations.

\subsection{Robustness Against Manually Designed Input Perturbations}
\label{sec:adversarial_atk}
Given the significant concerns with model robustness to random perturbations, this section presents an analysis {focusing specifically on} manually designed adversarial perturbations. Recent years have witnessed great advance in SOD driven by the progress of deep learning. However, whether the deep SOD models are as powerful as they seem is a question to worth pondering. Meanwhile, DNNs have been previously found to be susceptible to adversarial attacks, where visually imperceptible perturbations lead to completely different predictions~\cite{szegedy2013intriguing}. Though intensively studied in classification tasks, adversarial$_{\!}$ attacks$_{\!}$ in$_{\!}$ SOD$_{\!}$ are$_{\!}$ rarely$_{\!}$ explored. As SOD has been integrated as a critical part in many security systems and commercial projects, SOD models also have potential risks of being attacked. Specifically, SOD  plays a significant role in many security systems, for detecting the candidates of interest targets from remote sensing images~\cite{li2019nested}, video surveillance data~\cite{mehmood2015saliency}, or sensor signals of autonomous vehicles~\cite{zhang2016instance}. In such situation, examining the robustness of SOD models is rather important because the insecurity of SOD modules may cause severe losses, \eg, the criminals may use
inconspicuous adversarial perturbations to fool SOD modules and then cheat the surveillance systems.
Besides, SOD has benefited many commercial projects such as photo editing~\cite{wang2018deepcrop}, and image/video compression~\cite{guo2009novel}. The adversarial attacks launched by hackers on the embedded SOD modules would inevitably affect the functioning of commercial products and impacting users, causing losses for the developers and companies. Therefore, studying the robustness of SOD models is crucial for defending these applications against malicious attacks. In this section, we study the robustness against adversarial attacks and transferability of adversarial examples targeting different SOD models. Our observations are expected to shed light on adversarial attacks and defenses for SOD, providing a better understanding of vulnerabilities of deep SOD models and improving the robustness of SOD involved practical applications.
\begin{figure}[t]
  \centering
      \includegraphics[width=1 \linewidth]{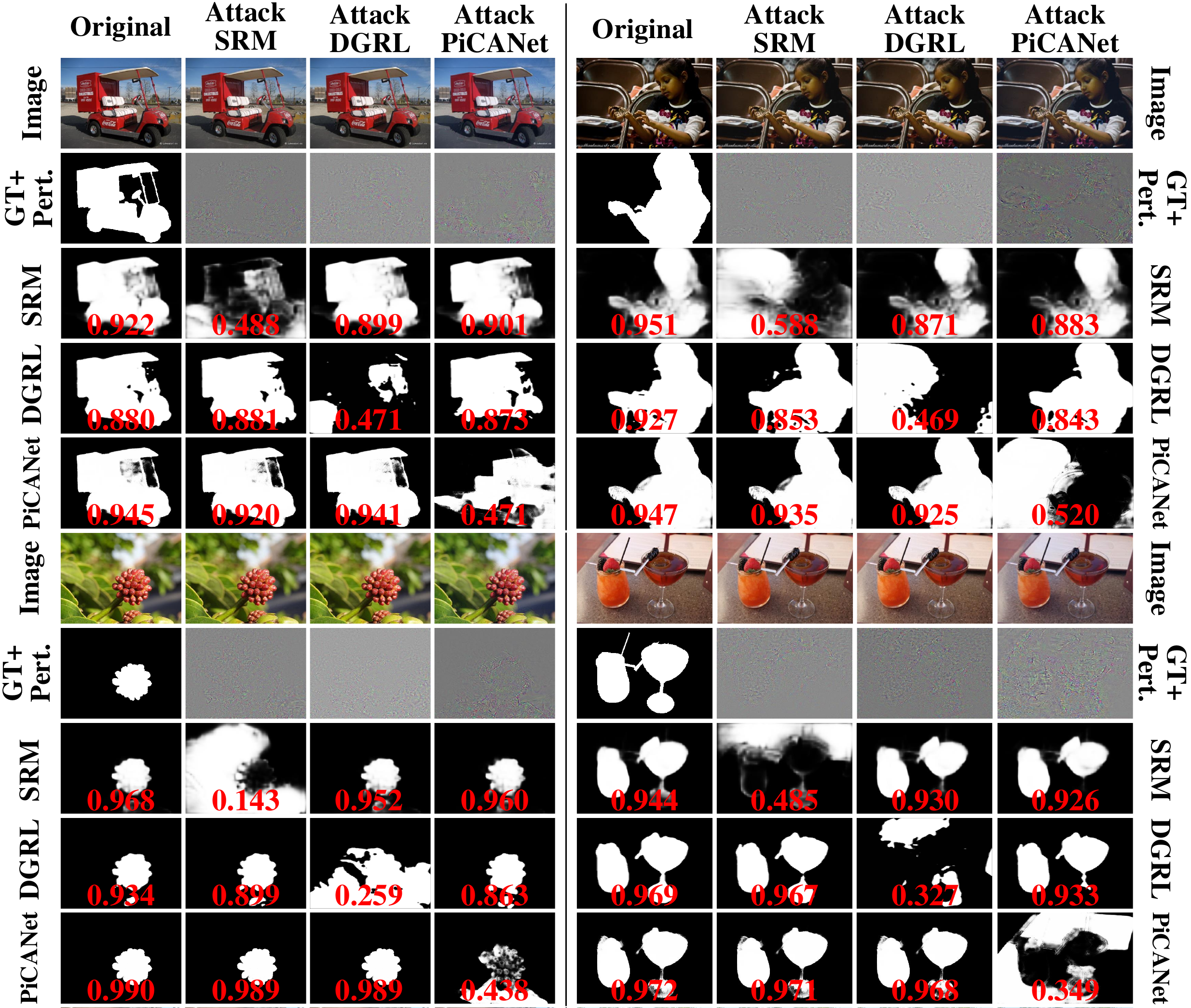}
\caption{Examples of SOD prediction under adversarial perturbations of different target networks. The
perturbations are magnified by $10$ for better visualization. {Red} for max~\texttt{F}.
See \S\ref{sec:adversarial_atk} for details.
}
\label{fig:Atk_examples}
\end{figure}

\subsubsection{Robustness of SOD Against Adversarial Attacks}
For measuring the robustness of deep SOD models, we adopt and modify an adversarial attack algorithm designed for semantic segmentation, \ie, {Dense Adversary Generation} (DAG)~\cite{xie2017adversarial}. We choose three representative deep models, \ie, SRM\!~\cite{Wang_2017_ICCV}, DGRL\!~\cite{wang2018detect} and PiCANet\!~\cite{liu2018picanet} for our study. The experiment is conducted on the hybrid benchmark introduced in \S\ref{sec:attribute_eval}.
{Following~\cite{xie2017adversarial}, we measure the perceptibility of the adversarial examples by computing the average perceptibility of the adversarial perturbations generated from the hybrid benchmark. The values for the three models are $3.54\!\times\!10^{-3}$, $3.57\!\times\! 10^{-3}$, and $3.51\!\times\!10^{-3}$, respectively.}

Exemplar adversarial cases are shown in Fig.~\ref{fig:Atk_examples}.
{As can be seen, the adversarial attacks can prevent the SOD models from producing reliable salient object candidates.} Quantitative results are listed in Table~\ref{table:adv_atk}. The underlined entries of Table~\ref{table:adv_atk} reveal that the three deep SOD models investigated are vulnerable to adversarial perturbations of the inputs. However, as can be observed by comparing Tables~\ref{table:input_p} and~\ref{table:adv_atk}, the models are more robust to random input perturbations.
These differences in robustness might be interpretated by the the distance from the inputs to the decision boundary in high dimensional space. The intentionally designed adversarial inputs often lie closer to the decision boundary than the random inputs~\cite{fawzi2016robustness}, and can thus more easily cause pixel-wise misclassification.

\subsubsection{Transferability Across Networks}
Previous research has revealed that adversarial perturbations can be transferred across networks, \ie adversarial examples targeting one model can mislead another without any modification~\cite{papernot2016transferability}. This transferability is widely used for black-box attacks against real-world systems. To investigate the transferability of perturbations for deep SOD models, we use the adversarial perturbation computed on one SOD model to attack another.

Table~\ref{table:adv_atk} shows the experimental results for the three models under investigation (SRM~\cite{Wang_2017_ICCV}, DGRL~\cite{wang2018detect} and PiCANet~\cite{liu2018picanet}). While the DAG attack leads to severe performance drops for the targeted model (see the diagonal), it causes much less degradation to other models, {\ie, the transferability between models of different network structures is weak for SOD task, which is similar to the transferability observed for semantic segmentation, as analyzed in~\cite{xie2017adversarial}}.
This may be because the gradient directions of different models are orthogonal to each other~\cite{liu2017delving}, so the gradient-based attack in the experiment transfers poorly to non-targeted models.
However, adversarial images generated from an ensemble of multiple models might generate non-targeted adversarial instances with better transferability~\cite{liu2017delving}, which would be a great threat to deep SOD models.

\begin{table}[t]
\centering
\caption{Results for adversarial attack experiments. Max~\texttt{F}$\uparrow$ on the \textbf{hybrid benchmark} is presented when exerting adversarial perturbations from different models. Worst results are \underline{\textbf{underline}}.  See \S\ref{sec:adversarial_atk} for details.}
\begin{threeparttable}
\resizebox{0.35\textwidth}{!}{
\setlength\tabcolsep{2pt}
\renewcommand\arraystretch{1.05}
\begin{tabular}{|r||c|c|c|}  
\hline\thickhline
Attack from
&~~~SRM~\cite{Wang_2017_ICCV}~~~ &~~DGRL~\cite{wang2018detect}~~  &PiCANet~\cite{liu2018picanet} \\
\hline
\hline
None &.817 &.831 &.848 \\
\hline
SRM~\cite{Wang_2017_ICCV} &\underline{\textbf{.263}} &.780 &.842 \\
\hline
DGRL~\cite{wang2018detect} &.778 &\underline{\textbf{.248}} &.844 \\
\hline
PiCANet~\cite{liu2018picanet} &.772 &.799 &\underline{\textbf{.253}} \\
\hline
\end{tabular}
}
\end{threeparttable}
\label{table:adv_atk}
\end{table}

\subsection{Cross-Dataset Generalization Evaluation}
\label{sec:cd_generalization}
Datasets are responsible for much of the
recent progress in SOD, not just as sources for training deep models, but also
as means for measuring and comparing performance.
Datasets are collected with the goal of representing the
visual world, {and to summarize the algorithm as a single number (\ie, benchmark score).}
A concern thus {arises}: it is necessary to evaluate how well a particular dataset
represents the real world; or, more specifically, to quantitatively measuring the dataset's generalization ability. Unfortunately, previous studies~\cite{borji2015salient} are quite limited -- mainly concerning the degrees of center bias in different SOD datasets. Here, we follow~\cite{Torralba2011Unbiased} to assess how general SOD datasets
are. We study the generalization and {difficulty} of several mainstream SOD datasets by performing a cross-dataset analysis, \ie, training on one dataset, and testing on the others. {We expect} our experiments to stimulate discussion in the community regarding this essential but largely neglected issue.

We first train a typical SOD model on one dataset, and then explore how well it generalizes to a representative set of other datasets, compared with its performance on the ``native'' test set.
Specifically, we implement the typical SOD model as a bottom-up/top-down structure, which has been the most standard and popular SOD architecture these years and is the basis of many current top-performing models~\cite{liu2019simple,su2019selectivity,zhao2019egnet,wu2019stacked}. As shown in Fig.~\ref{fig:SOD_network_cdg}, the encoder part is borrowed from VGG16~\cite{simonyan2014very}, and the decoder consists of three convolutional layers that gradually refine the saliency prediction. We pick six representative datasets~\!\cite{ChengPAMI,shi2015hierarchical,yang2013,li2015visual,wang2017,Fan_2018_ECCV}. For each dataset, we train the SOD model with $800$ randomly selected training images and test it on $200$ other validation images. Please note that {a total of} $1,000$ is the maximum possible number of images considering the size of the smallest selected dataset, ECSSD~\cite{shi2015hierarchical}.

\begin{figure}[t]
  \centering
      \includegraphics[width=1\linewidth]{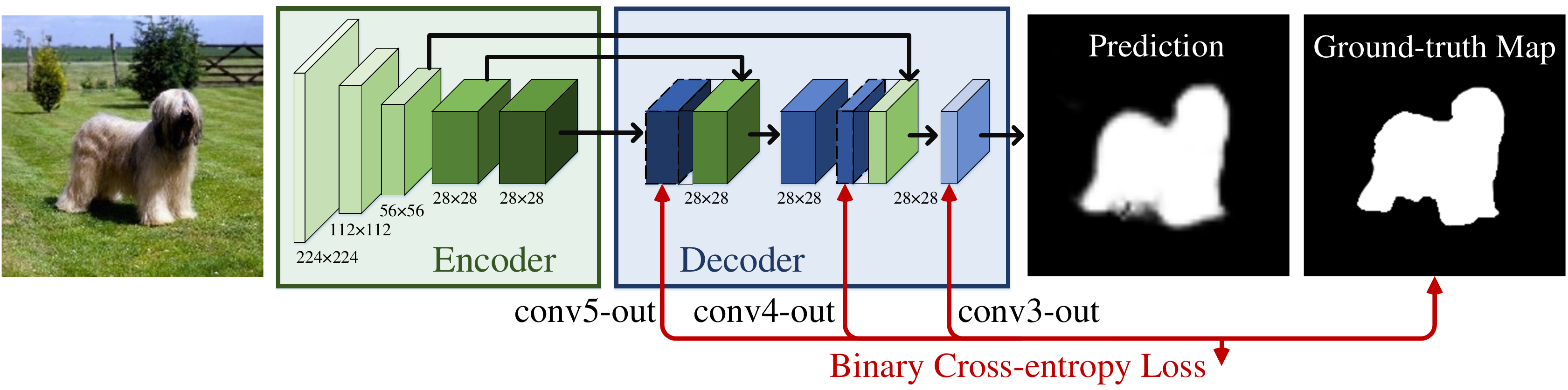}
\caption{Network architecture of the SOD model used in cross-dataset generalization evaluation. See \S\ref{sec:cd_generalization} for more detailed descriptions.}
\label{fig:SOD_network_cdg}
\end{figure}

\begin{table}[t]
\centering
\caption{Results for cross-dataset generalization experiment. Max~\texttt{F}$\uparrow$ for saliency prediction when training on one dataset (rows) and testing on another (columns).
``Self'' refers to training and testing on the same dataset (same as diagonal). ``Mean Others'' indicates average performance on all except self. See \S\ref{sec:cd_generalization} for details.}
\begin{threeparttable}
\resizebox{0.49\textwidth}{!}{
\setlength\tabcolsep{1pt}
\renewcommand\arraystretch{1.2}
\begin{tabular}{|r|cccccc|cc|c|}  
\hline\thickhline
{\diagbox[height=2.80em,width=8em,trim=l]{\!Train on:\!\!}{\!\!Test on:}}
&\tabincell{c}{MSRA-\\10K\cite{ChengPAMI}} &\tabincell{c}{ECSSD\\\cite{shi2015hierarchical}} &\tabincell{c}{DUT-OM\\RON\cite{yang2013}}  &\tabincell{c}{HKU-\\IS\cite{li2015visual}} &~\tabincell{c}{DUTS\\\cite{wang2017}}~ &~\tabincell{c}{SOC\\\cite{Fan_2018_ECCV}}~
&~Self &~\tabincell{c}{Mean\\others}&\tabincell{c}{Percent\\drop$\downarrow$} \\
\hline
MSRA10K\cite{ChengPAMI} &\textbf{.875} &.818 &.660 &.849 &.671 &.617   &.875 &.723 &17\% \\
\hline
ECSSD\cite{shi2015hierarchical}  &.844 &\textbf{.831} &.630 &.833 &.646 &.616   &.831 &.714 &14\% \\
\hline
DUT-OMRON\cite{yang2013} &.795 &.752 &\textbf{.673} &.779 &.623 &.567   &.673 &.703 &-5\%  \\
\hline
HKU-IS\cite{li2015visual} &.857 &.838 &.695 &\textbf{.880} &.719 &.639   &.880 &.750 &15\%  \\
\hline
DUTS\cite{wang2017} &.857 &.834 &.647 &.860 &\textbf{.665} &.654   &.665 &.770 &-16\%  \\
\hline
SOC\cite{Fan_2018_ECCV} &.700 &.670 &.517 &.666 &.514 &\textbf{.593}   &.593 &.613 &-3\%  \\
\hline
Mean others &.821 &.791 &.637 &.811 &.640 &.614   &- &- &-  \\
\hline
\end{tabular}
}
\end{threeparttable}
\label{table:cdg}
\end{table}

Table~\ref{table:cdg} summarizes the results of cross-dataset generalization, measured by max~\texttt{F}.
Each column corresponds to the
performance when training on all the datasets separately
and testing on one. Each row indicates training
on one dataset and testing on all of them.  Since our training/testing protocol is different from the one
used in the benchmarks mentioned in previous sections, the
actual performance numbers are not meaningful. Rather,
it is the relative performance difference that matters.
Not surprisingly, we observe that the best results are achieved
when training and testing on the same dataset. By looking at the
numbers across each column, we can determine how easy a dataset is for models trained on the other datasets. By looking at the numbers across one row, we can determine how good a dataset is at generalizing to the others. We find that SOC~\cite{Fan_2018_ECCV} is the most difficult dataset (lowest column, \textit{Mean others} $0.614$). MSRA10K~\cite{ChengPAMI} appears to be the easiest one (highest column, \textit{Mean others} $0.811$), and generalizes the worst (highest row, \textit{Percent drop} $17\%$). DUTS~\cite{wang2017} is shown to have the best generalization ability (lowest row, \textit{Percent drop} $-16\%$).

Based on these analyses, we would make the following recommendations for SOD datasets:
1) For training deep models, DUTS~\cite{wang2017} is a good choice because it has the best generalizability.
2) For testing, SOC~\cite{Fan_2018_ECCV} {is good for assessing} the worst-case performances, since it is the most challenging dataset. DUT-OMRON~\cite{yang2013} and DUTS-\textit{test}~\cite{wang2017} deserve more considerations as they are also very {difficult}.

\section{More Discussions}
\label{sec:discussions}
{Our previous systematic review and empirical studies characterized the models (\S\ref{sec:sod_models}), datasets (\S\ref{sec:soddatasets}), metrics (\S\ref{sec:evametric}), and challenges (\S\ref{sec:benchmarking}) of deep SOD. Here we further posit active research directions, and outline several open issues.}

\subsection{Model Design}\label{sec:model_design}
{Based on the review of deep SOD network architectures in \S\ref{sec:sod_network_structure}, as well as recent advances in related fields,} we here discuss several essential directions for SOD model design.

\noindent\textbullet~\textbf{Network topology.} Network topology determines the within-network information flow, which directly affects model capacity and training difficulty {and thus influences} the best possible performance.
To figure out an effective network topology for SOD, diverse architectures have been explored (\S\ref{sec:sod_network_structure}), such as multi-stream networks, side-out fusion networks, as well as bottom-up/top-down networks. However, all these network architectures are hand-designed. Thus, a promising direction would be to use \textit{automated machine learning} (AutoML) algorithms, such as \textit{neural architecture search}~\cite{zoph2017neural}, to automatically search for the best-performing SOD network topology.

\noindent\textbullet~\textbf{Loss function.} Most deep SOD methods are trained with the standard binary cross-entropy loss, which may fail to fully capture the quality factors for the SOD task. Only a few efforts have been made to derive losses from SOD evaluation metrics~\cite{Wang_2018_CVPR}. Thus, it is worth exploring more effective SOD loss functions, such as the mean intersection-over-union loss~\cite{berman2018lovasz} and affinity field matching loss~\cite{ke2018adaptive}.

\noindent\textbullet~\textbf{Adaptive computation.} Currently, all deep SOD models are fixed feed-forward structures. However, most parameters model high-level features that, in contrast to low-level and many mid-level concepts,
cannot be broadly shared across categories/scenes. {As such}, we would like to ask the following question: What if a SOD model could directly execute
certain layers that can best explain the saliency patterns in a given scene?
To answer this, one could leverage adaptive computation techniques~\cite{bengio2015conditional,Veit_2018_ECCV} to vary the amount of computation on-the-fly, \ie, by selectively activating part of the network in an input-dependent fashion. This could bring a better trade-off between network depth and computational
cost.  On the other hand, adapting inference pathways for different inputs would provide finer-grained discriminative ability for various attributes.
Therefore, exploring dynamic network structures in SOD is promising for improving both efficiency and effectiveness.

\subsection{Data Collection}
\label{sec:dataset_collection}
{Our previous {discussions} (\S\ref{sec:soddatasets}) and analyses (\S\ref{sec:attribute_eval} and \S\ref{sec:cd_generalization}) on current SOD datasets revealed several factors that are essential for {future} dataset collection.}

\noindent\textbullet~\textbf{Annotation inconsistency.} 
Though existing SOD datasets play a  critical role in training and evaluating modern SOD models, annotation inconsistencies among different SOD datasets {have essentially been ignored by} the community.
 The inconsistencies are mainly caused by separate subjects and rules/conditions during dataset annotation (see Fig.~\ref{fig:Anno_inconsist}). To ease annotation burdens, most current SOD datasets only have a few human annotators directly identify the salient objects, instead of considering real human eye fixation behavior.
Maintaining annotation consistency among newly collected datasets is {an important} consideration.

\begin{figure}[t]
  \centering
      \includegraphics[width=0.99 \linewidth]{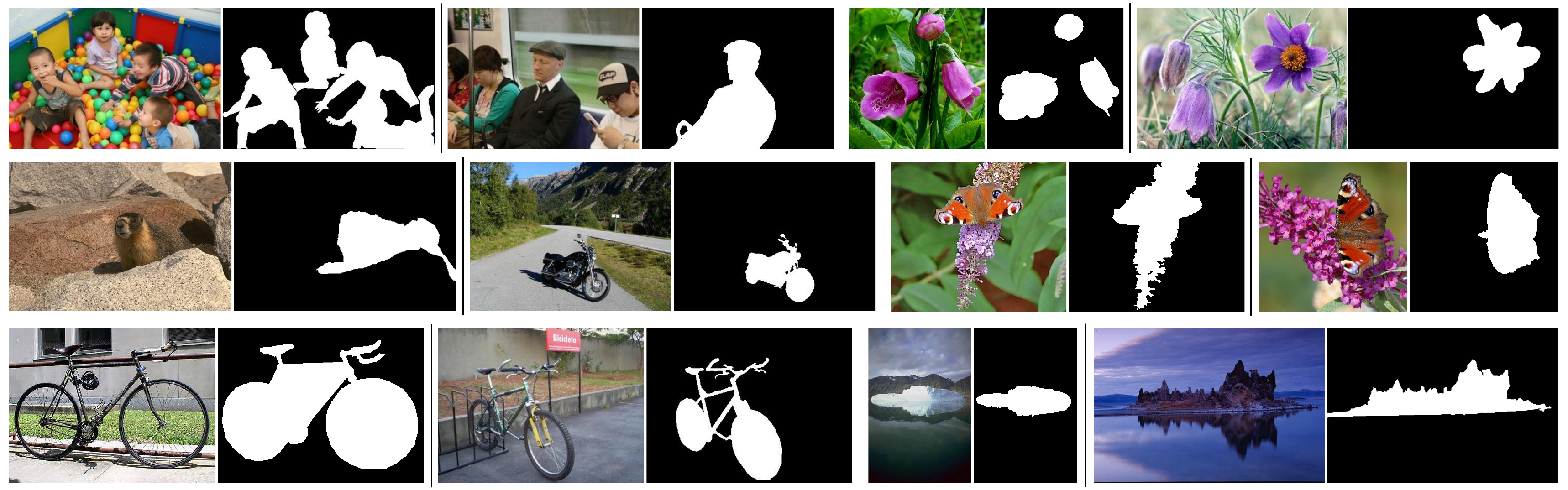}
\caption{Examples for annotation inconsistency. Each row shows two exemplar image pairs. See \S\ref{sec:dataset_collection} for more detailed descriptions.}
\label{fig:Anno_inconsist}
\end{figure}

\noindent\textbullet~\textbf{Coarse \textit{v.s.} fine annotation.}
Modern SOD datasets all have pixel-level annotations, which greatly boosts the performance of deep SOD models.
However, pixel-wise ground-truths are very costly to collect considering the complex object boundaries and the intense data requirement. {Further}, the annotation qualities of different datasets are different (see bicycles in Fig.~\ref{fig:Anno_inconsist}).
Finer labels are believed to be essential for high-quality saliency prediction, but usually {take} more time to collect.
Thus, given a limited budget, {finding} the optimal annotation strategy is an open problem.
Some works have studied the relationship between label quality and model performance in semantic segmentation~\cite{Zlateski_2018_CVPR}, {highlighting} a possible research direction for SOD dataset collection. In addition, current SOD models typically assume that the annotations are perfect. Thus, it {would also be} of value to explore robust SOD models that can learn saliency patterns from imperfectly annotated data.

\noindent\textbullet~\textbf{Domain-specific SOD datasets.}
SOD has shown potential in a wide range of applications, such as autonomous vehicles, video games, medical image processing, \etc. Due to the different visual appearances and semantic components, the saliency mechanisms in these applications are quite different from that of conventional natural images. Thus, collecting domain-specific datasets might benefit the application of SOD in certain scenarios, as observed in FP for crowds~\cite{jiang2014saliency}, webpages~\cite{zheng2018task} or driving~\cite{palazzi2017learning}, and better connect SOD to the biological top-down visual attention mechanism and human mental state.

\subsection{Saliency Ranking and Relative Saliency}\label{sec:relative_sal}
Current algorithms seems over-focused on directly regressing the saliency map to pursue a high benchmarking number, while neglecting the fact that the absolute magnitude of values in a saliency map
might be less important~than the relative saliency values among objects~\cite{li2014secrets}. Though the relative value/rank order is rarely considered in the context of benchmarking metrics ({with the exception of}~\cite{Islam_2018_CVPR}), it is crucial for better modeling human visual attention behavior. {This} is, in essence, a selection process that centers our attention on certain important elements of the surroundings, while blending other relatively unimportant things into the background. This not only hints at one shortcoming of existing benchmarking paradigms and data collection strategies, but also reveals a common limitation of current methods. Current {state-of-the-arts fall} short at {determining} the relative importance of objects, such as {identifying} the most important person in a crowded room. This is also evidenced by the experiments in \S\ref{sec:attribute_eval}, which show that deep models face great difficulties in complex ($\mathcal{CS}$), indoor (\textit{Indoor}) or multi-object ($\mathcal{MO}$) scenes. In other words, deep SOD models, though good at semantic modeling, require higher-level image understanding. Exploring more  powerful network designs that explicitly reason the relative saliency and revisiting classic cognitive theories are both promising directions to overcome this issue.

\subsection{{Linking} SOD to Visual Fixations}\label{sec:with_fx}
The strong correlation between eye movements (implicit saliency) and explicit object saliency has been explored {throughout} history~\cite{mishra2012active,masciocchi2009everyone,li2014secrets,borji2015salient,borji2015salientTip}. However, despite the deep connections between the problems of FP and SOD, the major computational models {of the two tasks remain largely distinct}; only a few SOD models consider {both} tasks simultaneously~\cite{chen2017look,Wang_2018_CVPR,kruthiventi2016saliency}. This is mainly due to the {overemphasis on the specific setting of SOD and the design bias of current SOD datasets}, which overlooks the connection to eye fixations during data annotation. As stated in~\cite{li2014secrets}, such dataset
design bias not only creates a {discomforting disconnection} between FP and SOD, but
also further misleads the algorithm designing. Exploring classic visual attention theories in SOD is a promising and crucial direction which could make SOD models more consistent with the visual processing of human visual system and {provide} better explainability. In addition, the ultimate goal of visual saliency modeling is to understand the underlying rationale of the visual attention mechanism. However, with the {current} focus on exploring more powerful neural network architectures and beating the latest benchmark numbers on {different} datasets, have we perhaps lost sight of the original purpose? The solution to these problems requires dense collaborations {between the} FP and SOD communities.

\subsection{{Learning} SOD in a {Weakly-/Unsupervised} Manner}\label{sec:with_unw}
Deep SOD methods are typically trained in a fully-supervised manner with a plethora of finely-annotated pixel-level ground-truths. However, it is highly costly and time-consuming to construct a large-scale, well-annotated SOD dataset. Though some efforts have been made {to achieve SOD with limited supervision}, \ie, {by} leveraging category-level labels~\cite{wang2017,cao2018lateral,li2019supervae} or pseudo pixel-wise annotations~\cite{zhang2017supervision,li2018weakly,zhang2018deep,li2018contour,zeng2019multi}, there is still a notable gap with the fully-supervised counterparts. In {contrast}, humans usually learn with little or even no supervision. {Since} the ultimate goal of visual saliency modeling is to understand the visual attention mechanism, learning SOD in an {weakly-/unsupervised} manner would be of great value to both the research community and real-world applications. {Further, it would also help us} understand which factors truly drive our attention mechanism and saliency pattern understanding. Given the massive number of algorithmic breakthroughs over the past few years, we can expect a
flurry of innovation towards this promising direction.

\subsection{{Pre-training with Self-Supervised Visual Features}}
Current deep SOD methods are typically built on ImageNet-pretrained networks, and fine-tuned on SOD datasets. It is believed that parameters trained on ImageNet can serve as a good starting point to accelerate the convergence of training and {prevent} overfitting on smaller-scale SOD datasets.
Besides pre-training deep SOD models on the \textit{de facto} dataset, ImageNet, another {option} is to leverage self-supervised learning techniques~\cite{jing2020self} to learn effective visual features from a vast amount of unlabeled images/videos.
The visual features can be learned through various pretext tasks like image inpainting~\cite{pathak2016context}, colorization~\cite{larsson2017colorization}, clustering~\cite{caron2018deep}, \etc, and can be generalized to other vision tasks.
Fine-tuning the SOD models on parameters trained from self-supervised learning is promising to yield better performance compared to the ImageNet initialization.

\subsection{Efficient SOD for Real-World Application}\label{sec:real_world}
Current top-leading deep SOD models are designed to be complicated in order to achieve increased learning capacity and improved performance. However, more ingenuous and light-weight architectures are required to fulfill the requirements of mobile and embedded applications, such as robotics, autonomous driving, augmented reality, \etc. The degradation of accuracy and generalization ability caused by model scale deduction {should be minimal}. To facilitate the application of SOD in real-world scenarios, it is {possible} to utilize model compression~\cite{bucilua2006model} or knowledge distillation~\cite{hinton2015distilling,romero2014fitnets} techniques to develop compact and fast SOD models with competitive performance.
Such compression techniques have already been shown effective in improving generalization ability and alleviating under-fitting for training efficient object detection models~\cite{chen2017learning}.

\section{Conclusion}
\label{sec:conclusion}

In this paper we present, to the best of our knowledge, the first comprehensive review of SOD {focusing} on deep learning techniques.
We first provide novel testimonies for categorizing deep SOD models from several distinct perspectives, including network architecture, level of supervision, \etc.
We then cover the contemporary literature on popular SOD datasets and evaluation criteria, providing a thorough performance benchmarking of major SOD methods and offering recommendations for several datasets and metrics {that can be used} to consistently assess {different} models. Next, we consider several previously under-explored issues {related} to benchmarking and baselines.
In particular, we study the strengths and weaknesses of deep and non-deep SOD models by compiling and annotating a new dataset and evaluating several representative models on it, {revealing} promising directions for future efforts.
We also study the robustness of SOD methods by analyzing the effects of various perturbations on the final performance.
Moreover, for the first time in the field, we investigate the robustness of deep SOD models {to} maliciously designed adversarial perturbations and the transferability of these adversarial examples, {providing} baselines for future research.
In addition, we analyze the generalization and difficulty of existing SOD datasets through a cross-dataset generalization study, and quantitatively reveal the dataset bias.
We finally {introduce} several open issues and challenges of SOD in the deep learning era, providing insightful discussions and identifying a number of potentially fruitful directions forward.

In conclusion, SOD has achieved notable progress thanks to the striking development of deep learning techniques. {However,} there are still under-explored problems on achieving more efficient {model designs}, training, and {inference} for both academic research and real-world applications.
We expect this survey to provide an effective way to understand {current} state-of-the-arts and, more importantly, insight for the future exploration of SOD.

\ifCLASSOPTIONcaptionsoff
  \newpage
\fi



%
{\small
\bibliographystyle{IEEEtran}

\bibliography{egbib}
}

\vspace{-25pt}
\begin{IEEEbiographynophoto} {Wenguan Wang} received his Ph.D. degree from Beijing Institute of Technology in 2018. He is
currently a postdoc scholar at ETH Zurich, Switzerland.
From 2016 to 2018, he was a visiting Ph.D. student in University of California, Los Angeles. From 2018 to 2019, he was a senior scientist at Inception
Institute of Artificial Intelligence, UAE. His current research interests include computer vision and deep learning.
\end{IEEEbiographynophoto}
\vspace{-25pt}
\begin{IEEEbiographynophoto} {Qiuxia Lai} received the B.E. and M.S. degrees in the School of Automation from Huazhong University of Science and Technology in 2013 and 2016, respectively. She is currently pursuing the Ph.D. degree in The Chinese University of Hong Kong. Her research interests include image/video processing and deep learning.
\end{IEEEbiographynophoto}
\vspace{-25pt}
\begin{IEEEbiographynophoto} {Huazhu Fu} (SM'18) received the Ph.D. degree from Tianjin University, China, in 2013. He was a Research Fellow with Nanyang Technological University, Singapore for two years. From 2015 to 2018, he was a Research Scientist with the Institute for Infocomm Research, Agency for Science, Technology and Research, Singapore. He is currently a Senior Scientist with Inception Institute of Artificial Intelligence, UAE. His research interests include computer vision and medical image analysis. He is an Associate Editor of IEEE TMI and IEEE Access.
\end{IEEEbiographynophoto}
\vspace{-25pt}
\begin{IEEEbiographynophoto} {Jianbing Shen} (M'11-SM'12) is a Professor with the School of Computer Science, Beijing Institute of Technology. He has published about 100 journal and conference papers such as \textit{TPAMI}, \textit{CVPR}, and \textit{ICCV}. He has obtained many honors including the Fok Ying Tung Education Foundation from Ministry of Education, the Program for Beijing Excellent Youth Talents from Beijing Municipal Education Commission, and the Program for New Century Excellent Talents from Ministry of Education. His research interests include computer vision and deep learning.  He is an Associate Editor of IEEE TNNLS, IEEE TIP and Neurocomputing.
\end{IEEEbiographynophoto}
\vspace{-25pt}
\begin{IEEEbiographynophoto} {Haibin Ling} received the PhD degree from University of Maryland in 2006. From 2000 to 2001, he was an assistant researcher at Microsoft Research Asia. From 2006 to 2007, he worked as a postdoc at University of California Los Angeles. After that, he joined Siemens Corporate Research as a research scientist. Since 2008, he has been with Temple University where he is now an Associate Professor. He received the Best Student Paper Award at the ACM UIST in 2003, and the NSF CAREER Award in 2014. He is an Associate Editor of IEEE TPAMI, PR, and CVIU, and served as Area Chairs for CVPR 2014, 2016 and 2019.
\end{IEEEbiographynophoto}
\vspace{-25pt}
\begin{IEEEbiographynophoto}
{Ruigang Yang} is currently a full professor of Computer Science at the University of Kentucky.
His research interests span over computer vision and computer graphics, in particular in 3D reconstruction and 3D data analysis.
He has received a number of awards, including the US National Science Foundation Faculty Early Career Development (CAREER) Program Award in 2004, and the best Demonstration Award at CVPR 2007.
He is currently an associate editor of IEEE TPAMI.
\end{IEEEbiographynophoto}
\vspace{-20pt}



\vfill


\end{document}